\documentclass{article}

\usepackage{amsmath,amsfonts,bm}

\def\eqref#1{equation~\ref{#1}}

\def\1{\bm{1}}

\DeclareMathAlphabet{\mathsfit}{\encodingdefault}{\sfdefault}{m}{sl}
\SetMathAlphabet{\mathsfit}{bold}{\encodingdefault}{\sfdefault}{bx}{n}

\DeclareMathOperator*{\argmax}{arg\,max}
\DeclareMathOperator*{\argmin}{arg\,min}

\usepackage{microtype}
\usepackage{graphicx}
\usepackage{booktabs} 

\usepackage{bbm}
\usepackage{subcaption}
\usepackage{amsmath}
\usepackage{amssymb}
\usepackage{cancel}
\usepackage{xcolor}
\usepackage{algpseudocode}
\usepackage{float}
\usepackage{wrapfig}
\usepackage{booktabs}

\usepackage{hyperref}

\newcommand\bs[1]{\boldsymbol{#1}}
\newcommand\bv[1]{\mathbf{#1}}

\newcommand{\fedmix}{\texttt{FedMix}}
\newcommand{\fedavg}{\texttt{FedAvg}}

\usepackage[accepted]{icml2021}

\icmltitlerunning{Federated Mixture of Experts}

\begin{document}

\twocolumn[
\icmltitle{Federated Mixture of Experts}



\icmlsetsymbol{equal}{*}

\begin{icmlauthorlist}
\icmlauthor{Matthias Reisser}{quva,qc}
\icmlauthor{Christos Louizos}{qc}
\icmlauthor{Efstratios Gavves}{quva}
\icmlauthor{Max Welling}{quva,qc}

\end{icmlauthorlist}

\icmlaffiliation{quva}{QUVA Lab, University of Amsterdam}
\icmlaffiliation{qc}{Qualcomm AI Research}

\icmlcorrespondingauthor{Matthias Reisser}{mreisser@qti.qualcomm.com}

\icmlkeywords{Machine Learning, ICML}

\vskip 0.3in
]



\printAffiliationsAndNotice{}  

\begin{abstract}
Federated learning (FL) has emerged as the predominant approach for collaborative training of neural network models across multiple users, without the need to gather the data at a central location. One of the important challenges in this setting is data heterogeneity, \textit{i.e.} different users have different data characteristics. For this reason, training and using a single global model might be suboptimal when considering the performance of each of the individual user's data. In this work, we tackle this problem via Federated Mixture of Experts, \texttt{FedMix}, a framework that allows us to train an ensemble of specialized models.~\texttt{FedMix} adaptively selects and trains a user-specific selection of the ensemble members. We show that users with similar data characteristics select the same members and therefore share statistical strength while mitigating the effect of non-i.i.d data. Empirically, we show through an extensive experimental evaluation that \texttt{FedMix} improves performance compared to using a single global model across a variety of different sources of non-i.i.d.-ness. 
\end{abstract}

\section{Introduction}

\begin{figure}[t]
    \centering
    \includegraphics[trim=0.5cm 0.5cm 0.7cm 0cm, width=0.5\textwidth]{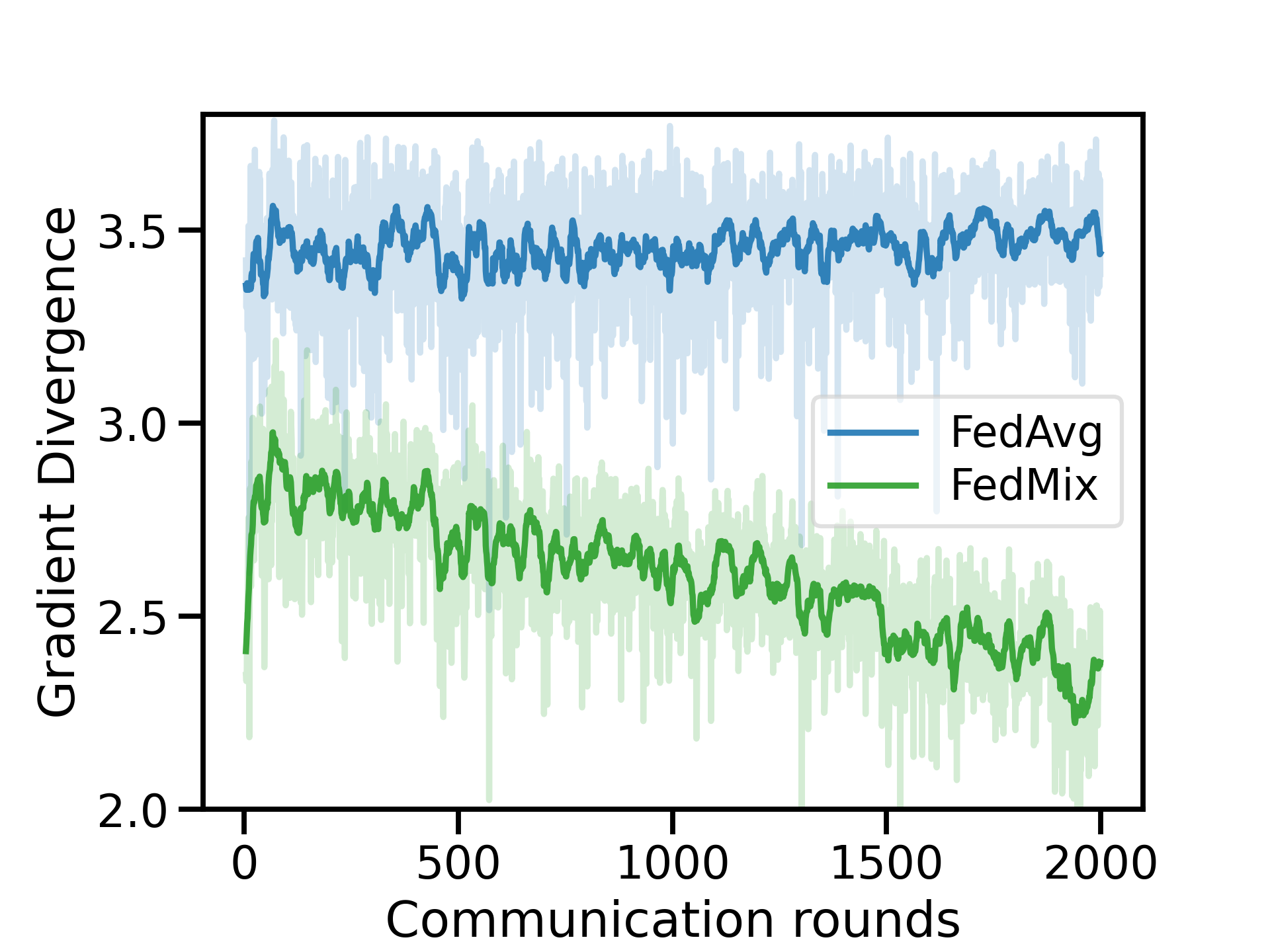}
    \caption{A sliding window of the gradient divergence (defined in Appendix \ref{sec:grad_alignment}), on Cifar10 in the setup of Section \ref{sec:experiments} for \fedavg{} and \fedmix{} ($K=4$).}
    \vspace{-20pt}
    \label{fig:GradDivergence}
\end{figure}

An ever-increasing amount of devices are being connected to the internet, sensing their environment, and generating vast amounts of data. The term federated learning (FL) has been established to describe the scenario where we aim to learn from the data generated by this ``federation'' of devices \citep{mcmahan2016communication}. Not only does the number of sensing devices increase, but also their processing power is increasing continuously to the point that it becomes viable to perform inference and training of machine learning models on device. In federated learning, the goal is to learn from these client devices' data without collecting the data centrally, which naturally allows for more private exchange of information.

Several challenges arise in the federated scenario. Federated devices are generally resource-constrained, both in their computational capacity as well as in communication bandwidth and latency. In a practical example, a smartphone has limited heat dissipation capacity and must communicate via Wi-Fi. From a global perspective, devices' processing power and network connection can be highly heterogeneous across geographical regions and socio-economical status of device owners, causing practical issues \citep{bonawitz2019towards} and raising questions of fairness in FL \citep{li2019fair, mohri2019agnostic}. Apart from this cross-device setting, challenges in the so-called cross-silo setting focus on privacy concerns, while computation and communication constrains move to the background \citep{kairouz2019advances}. 

One of the key challenges in FL that we aim to address in this work is the non-i.i.d. nature of the shards of data that are distributed across devices. In non-federated machine learning, assuming independent and identically distributed data is generally justifiable and not detrimental to model performance. In FL however, each client performs a series of parameter updates on its own data shard to amortize the costs of communication.
Over time, the direction of progress across shards with non-i.i.d. data starts diverging (as shown in Figure \ref{fig:GradDivergence}), which can set back training progress, significantly slow down convergence and decrease model performance \citep{hsu2019measuring}. 

To this end, we propose Federated Mixtnure of Experts (\fedmix{}), an algorithm for FL that allows for training an ensemble of specialized models instead of a single global model. In \fedmix{}, expert models are learning to specialize in regions of the input space such that, for a given expert, each client's progress on that expert is aligned. \fedmix{} allows each client to learn which experts are relevant for its shard and we show how it can be extended for inference on a previously unseen client. \fedmix{} shows competitive performance against the established standard in FL, \fedavg{} \citep{mcmahan2016communication, deng2020adaptive} across a range of visual classification tasks.

\section{Federated Mixture of Experts}

Federated learning~\citep{mcmahan2016communication} deals with the problem of learning a server model with parameters $\bv{w}$, \emph{e.g.}, a neural network, from a dataset of $N$ datapoints $\mathcal{D} = \{(\bv{x}_1, y_1),\dots, (\bv{x}_N, y_N)\}$  that is distributed across $S$ shards, \emph{i.e.}, $\mathcal{D} = \mathcal{D}_1 \cup \dots \cup \mathcal{D}_S$, \emph{without} accessing the shard-specific datasets directly. By defining a loss function $\mathcal{L}_s(\mathcal{D}_s; \bv{w})$ per shard, the total risk can be written as
\begin{align}
    &\argmin_\bv{w} \sum_{s=1}^S \frac{N_s}{N} \mathcal{L}_s(\mathcal{D}_s; \bv{w}),\\ &\mathcal{L}_s(\mathcal{D}_s; \bv{w}) := \frac{1}{N_s}\sum_{i=1}^{N_s} L(\mathcal{D}_{si}; \bv{w}).
\end{align}
It is easy to see that this objective corresponds to empirical risk minimization over the joint dataset $\mathcal{D}$ with a loss $L(\cdot)$ for each datapoint. In federated learning one is interested in reducing the communication costs; for this reason~\cite{mcmahan2016communication} propose to do multiple gradient updates for $\bv{w}$ in the inner optimization objective for each shard $s$, thus obtaining ``local'' models with parameters $\bv{w}_s$. These multiple gradient updates are denoted as ``local epochs'', \emph{i.e.}, amount of passes through the entire local dataset, with an abbreviation of $E$. Each of the shards then communicates the local model $\bv{w}_s$ to the server and the server updates the global model at ``round'' $t$ by averaging the parameters of the local models $\bv{w}^t = \sum_s \frac{N_s}{N}\bv{w}^t_{s}$. This constitutes federated averaging (\texttt{FedAvg})~\citep{mcmahan2016communication}, the standard in federated learning.

One of the main challenges in federated learning is the fact that usually the data are non-i.i.d. distributed across the shards $S$, that is $p(\mathcal{D}|s_i) \neq p(\mathcal{D}|s_j)$ for $i \neq j$. On the one hand, this can make learning a single global model from all of the data with the classical \fedavg{} problematic. On the other hand, there is one extreme that does not suffer from this issue; learning $S$ individual models, \emph{i.e.}, only optimizing $\bv{w}_s$ on $\mathcal{D}_s$. Although these individual models by definition do not suffer from non-i.i.d. data, clearly we should aim to do better and exchange meaningful information between clients to learn more robust and expressive models.

With \fedmix{}, we propose to strike a balance between the two aforementioned extremes; learning a single global model and learning $S$ individual models. For this reason, we revisit an old model formulation, the Mixture of Experts (MoE).
The classical formulation of a MoE model \citep{jacobs1991adaptive,jordan1994hierarchical} contains a set of $K$ experts and a gating mechanism that is responsible for choosing an expert for a given data-point. A MoE model for a data point $(\bv{x},y)$ can generally be described by
\begin{align}\label{eq:MoEStandard}
    p_{\bv{w}_{1:K}, \theta}(y|\bv{x}) = \sum_{z=1}^K p_{\bv{w}_z}(y|\bv{x},z)p_\theta(z|\bv{x}),
\end{align}
where $z$ is a categorical variable that denotes the expert, $\bv{w}_k$ are the parameters of expert $k$ and $\theta$ are the parameters of the selection mechanism.

The MoE was proposed as a model for datasets where different subsets of the data exhibit different relationships between input $\bv{x}$ and output $y$. Instead of training a single global model to fit this relationship everywhere, each expert performs well on a different subset of the input space. The gating function models the decision boundary between input regions, assigning data-points from subsets of the input region to their respective experts.

In this work, we show that, in the federated scenario, sub-dividing the input region through a MoE can alleviate the consequences of non-i.i.d. data by aligning gradient updates across experts (Figure \ref{fig:GradDivergence}).

In Federated Mixture of Experts (\fedmix{}) we enrich this model by conditioning the gating mechanism on the shard assignment $s$. Whatever characteristics make shard $s$ different from other shards can manifest in learning a different, localized gating mechanism that does not need to be communicated to the server. In choosing $K=1$, \fedmix{} recovers the standard setting of federated averaging. $K=S$ in combination with fixing $p(z=s|\bv{x}, s) = 1$ recovers $S$ independent models.
From a global perspective, we are interested in maximizing the following single objective:
\begin{align}
    &\sum_{s=1}^S \sum_{i=1}^{N_s} \log p_{\bv{w}_{1:K}, \theta_s}(y_{s,i}|\bv{x}_{s,i},s) = \nonumber\\&\sum_{s=1}^S \sum_{i=1}^{N_s} \log \Big[\sum_{z=1}^K p_{\theta_s}(z|\bv{x}_{s,i},s)p_{\bv{w}_z}(y_{s,i}|\bv{x}_{s,i},z) \Big] 
    \label{eq:globalObjective}
\end{align} 

\pagebreak
While it is possible to optimize Eq.~\ref{eq:globalObjective} directly, we have found empirically that it is hard to achieve both: avoiding collapse to a single expert, thus obtaining \fedavg{}, and specialization of the experts. Instead, we propose to form a variational lower-bound on Eq.~\ref{eq:globalObjective} with a global variational approximation $q_\phi(z|\dots)$ to the true posterior $p(z|\bv{x},y,s)$ with parameters $\bs{\phi}$. At test time, $p(y|\bv{x}^*,s)=\sum_{k=1}^K p(y|\bv{x}^*,z) p(z|\bv{x}^*,s)$ can be readily evaluated without requiring $q$. This allows us to condition $q_\phi(z|\dots)$ on any available side-information \textit{at training time} that might result in better specialization in the non-i.i.d. federated scenario. In this paper, we discuss several sources of non-i.i.d.-ness: label skew, \textit{i.e.}, different distributions $p(y|s)$ per shard, input-transformations, \textit{i.e.}, different $p(x|s)$, and different mappings $p(y|x,s)$, such as label permutations. 

In practice, other or additional known sources of misalignment could be included to further improve this approximation, such as a manufacturer-id for a medical device in a medical scenario, a geographic identifier, or general domain-specific information. We show a variety of artificial scenarios in the experimental section. For exposition in this paper, we always use $q_\phi(z|y)$, however the formulae and algorithms are equally applicable to other side-information. The lower bound to be maximized in \fedmix{} therefore is as follows:

\begin{align}\label{eq:LowerBound}
&\!\sum_{s=1}^S \sum_{i=1}^{N_s} \log p_{\bv{w}_{1:K}, \theta_s}(y_{s,i}|\bv{x}_{s,i},s) \geq \sum_{s=1}^S \sum_{i=1}^{N_s}\mathbb{E}_{q_\phi(z|y_{s,i})}[\nonumber \\ 
&\!\log p_{\bv{w}_z}(y_{s,i}|\bv{x}_{s,i},z)p_{\theta_s}(z|\bv{x}_{s,i},s)]\!+\!\beta H(q_\phi(z|y_{s,i})),
\end{align}

where $\beta$ is a hyperparameter that controls the entropy of the approximate posterior distribution; a low $\beta$ will result in $q_\phi(z|y)$ that are more concentrated around their most probable value whereas higher values will encourage more uncertain distributions. While the parameters in Eq.~\ref{eq:LowerBound} can be optimized via gradient descent, a closed-form update for the parameters $\phi$ based on Lagrange multipliers exists if each conditional is parameterized directly, \textit{i.e.} $\phi_c=q(z|y=c) \in [0,1]^K$. The solution for the probabilities of each expert $k$ conditioned on a given category $c$ becomes
\begin{align}
    &\phi_{c,k} = \frac{\left(\prod_{i=1}^{N_{s,c}} p(y_{s,i},z\!=\!k|\bv{x}_{s,i}, s)\!\right)^{\frac{1}{N_{s,c}\beta}}}{\sum_{z=1}^K \left(\prod_{i=1}^{N_{s,c}} p(y_{s,i},z\!=\!k|\bv{x}_{s,i}, s)\!\right)^{\frac{1}{N_{s,c}\beta}}}\\
    &\!\!\!= \!\frac{\exp\!\left(\frac{1}{N_{s,c}\beta}\sum_{i=1}^{N_{s,c}} \log p(y_{s,i},z\!=\!k|\bv{x}_{s,i}, s)\!\right)}{\sum_{z=1}^K \exp\!\left(\frac{1}{N_{s,c}\beta}\sum_{i=1}^{N_{s,c}} \log p(y_{s,i},z\!=\!k|\bv{x}_{s,i}, s)\!\right)},
\end{align} 
\textit{i.e.}, a softmax where the logits correspond to the average of the log-joint probabilities of the datapoints that belong to class $y = c$ and the prior probabilities of expert $k$ in shard $s$. We have omitted the dependence on parameters $\bv{w}_z,\theta_s$ for clarity. The full derivation is given in Appendix \ref{sec:fixedformmath}. It is interesting to see that the entropy hyperparameter $\beta$ acts as a ``temperature'' for the softmax, thus directly encouraging low or high entropy solutions for $q_\phi(z|y)$. This is in-line with our prior discussion about $\beta$.
In addition to the closed-form solution, such a parameterization is efficient w.r.t. communication overhead. 
Conditioning $q_\phi(z|y)$ on $s$ is possible and results in localized approximations with parameters $\phi_s$ that do not need to be communicated, however we found a global approximation to help align the gating mechanisms across shards.

\begin{figure}[h]
    
    \centering
    \includegraphics[trim=0cm 0cm 0cm 0cm,width=0.9\linewidth]{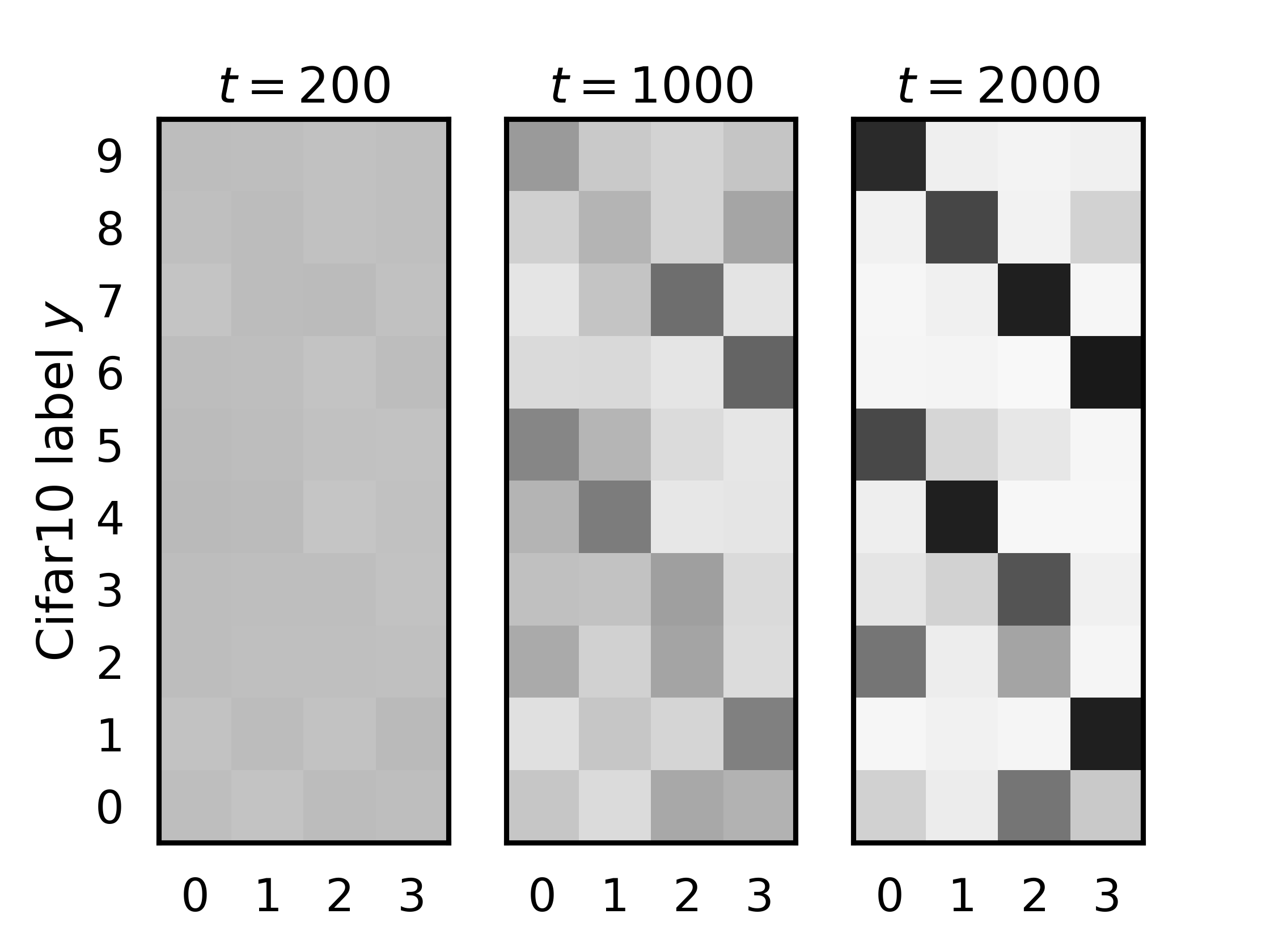}
    \caption{Visualization of $q_\phi(z|y)$ at different communication rounds $t$ for \fedmix{} with $K=4$ on Cifar10. Greyscale corresponds to probabilities; white corresponds to zero and black corresponds to one. Probabilities sum to one across experts (horizontally).}
    \label{fig:q_specialisation}

\end{figure}

\paragraph{Expert Specialization} 
Specialization of the experts is a key ingredient for \fedmix{} to be successful; with specialization, the gradients for each expert become aligned across shards (see Figure~\ref{fig:GradDivergence}) and the (training set) performance in general improves.

Nevertheless, we find a fundamental trade-off in the MoE formulation: highly specialized experts are useful only if the local gating networks make correct selections, which is not always the case for test data. Therefore, a successful application of the MoE formulation requires striking a balance between the specialisation of the experts and their robustness to being wrongly selected by the routing mechanism. During training, experts initially receive gradients from all data-points until $q_\phi(z|y)$ concentrates and enforces specialisation. We can thus control the speed of specialisation by tuning $\beta$ and performing dampening on the update of the probabilities of $q_\phi(z|y)$. We empirically find that this leads to experts that, while they do not specialize as aggressively, can provide reasonable predictions even on datapoints outside of their ``expertise'', thus improving the overall performance of \fedmix{}.

A possible drawback of specialization is that sometimes \fedmix{} prematurely completely prunes experts, \emph{i.e.}, $p_{\theta_s}(z=k| \bv{x}, s) \approx 0 \enskip \forall \bv{x}, s$. This can be undesirable as we lose model capacity that can be used for better modeling the data. As $q_{\phi}(z|y)$ is one of the main training signals of $p_{\theta_s}(z|\bv{x}, s)$, we introduce the marginal entropy term in the server, $H(E_{p(y)}[q_\phi(z|y)])$, as a regularizer that encourages using all of the experts. 
Figure \ref{fig:q_specialisation} show how $q_\phi(z|y)$ converges over time towards specialization of experts.

\paragraph{Personalization} Each of the specialized expert models that \fedmix{} provides can be thought as containing information about the specific subset of the clients that selects each particular expert. As a result, these experts can serve as a better starting point for personalization according to the data on a specific device. Personalization can be achieved by simply finetuning the models obtained from the server, \textit{i.e.}, $\bv{w}_{1:K}$, on the client specific training set $\mathcal{D}_{s,\text{train}}$ thus obtaining $\bv{w}^s_{1:K}$. Under the assumption that $p(\mathcal{D}_{\text{train}} | s) \approx p(\mathcal{D}_{\text{test}} | s)$, which is not unreasonable in the federated learning scenario where each device has its own data generating mechanism, such personalized models can then have better prediction capabilities on the test data of that device. Finetuning is performed by optimizing Eq.~\ref{eq:LowerBound} for a small number of steps (\textit{e.g.}, $E=1$) with respect to $\bv{w}_{1:K}, \phi$ and $\theta_s$. It should be mentioned that this finetuning procedure is not limited to \fedmix{} and can be performed on any federated learning method that involves global parameters shared by all of the clients, such as \fedavg{}.

During our experiments, we measure the performance of \fedmix{} by measuring the average local accuracy of these client specific personalized models on their respective client specific test sets. In order to avoid the extra finetuning step for each round, we use the last version of the model that each client communicated to the server as the personalized model. For fair comparisons, we employ the same finetuning procedure for all of our other baselines as well. 

\paragraph{Server Side Updates}
In a general federated learning algorithm, a central server selects a subset $S'\subset \{1,\dots,S\}$ of clients at time $t$ and transmits the current estimate of the global parameters $\bv{w}^t$ to them. These clients perform a series of mini-batch gradient updates with data from their shard $\mathcal{D}_s$ on a local loss function, which can come at the price of each client moving in possibly different directions in parameter space. In generalized \fedavg{} \citep{reddi2020adaptive}, the server interprets $\Delta_s^t = \bv{w}^t - \bv{w}_s^{t+1}$ as a single-step gradient update from client $s$, averages those gradients and applies an optimizer such as Adam \citep{kingma2014adam} to receive $\bv{w}^{t+1}$. In light of non-i.i.d. data across clients, this averaging strategy can result in slow progress since averaging updates in a highly non-convex parameter space can be sub-optimal. In \fedmix{}, this effect is mitigated since for a given expert, the data that is used to update its parameters are aligned better across shards.

\fedmix{} offers a second way to improve convergence speed by modifying the server-side updates. In generalized \fedavg{}, the individual gradients returned by the subset $S'$ of clients are averaged according to 
\begin{align}
    \Delta^t = \sum_{s=1}^{S'} p(s) \cdot \Delta_s^t\text{ ,}\quad p(s) = \frac{N_s}{N_{S'}}.
\end{align}
In \fedmix{}, we can speed up convergence by considering expert-specific updates $\Delta_{k,s}^t=\bv{w}_k^t - \bv{w}_{k,s}^{t+1}$. If a client $s$ pruned away expert $k$ from its local gating mechanism, $\Delta_{k,s}^t$ will be zero. We propose to normalize the effective magnitude of the resulting update $\Delta_{k}$ by up-weighing the updates of all other clients that do consider expert $k$ for their local mixture:
\begin{align}\label{eq:fedmixPavg}
    \!\!\!\Delta_k^t\!=\!\!\sum_{s=1}^{S'} p(s|z\!=\!k)\!\cdot\!\Delta_{k,s}^t\,\text{, } p(s|z\!=\!k) \!\propto\! p(z\!=\!k|s)p(s)
\end{align}
Computing $p(z|s) = \mathbb{E}_{\bv{x} \sim \mathcal{D}_s}[p_{\theta_s}(z|s,\bv{x})]$ prior to sending updates to the server involves evaluating potentially large neural network models which might not be desirable, depending on the situation and size of the local dataset. Therefore we approximate $p(z|s) \approx q_\phi(z|s) = \mathbb{E}_{y\sim \mathcal{D}_s}[q_\phi(z|y)]$, which involves just a single matrix multiplication.

\paragraph{Privacy implications}
The update rule described in Eq.~\ref{eq:fedmixPavg} requires access to the marginal $q(z|s)=\sum_y p(y|s)q_\phi (z|y)$ at the server. At the same time, the server has access to the parameters $\phi$ that were used in computing $p(z|s)$ before being sent to the server. Therefore, in principle, it could solve $q(z|s)=\sum_y p(y|s)q_\phi(z|y)$ with respect to $p(y|s)$ and thus obtain the marginal label distribution at the client. In practice this is not as straightforward to do as the probability matrix $q_\phi(z|y)$ is not always invertible and solutions that use the pseudo-inverse, empirically, are not very accurate in capturing the entire distribution. With the additional constraints that the marginal needs to sum to one, contains only positive elements and that $N_s \cdot p(y|s) \in \mathbb{Z}$, in some cases, a reconstruction can become possible. As the number of classes exceeds the number of experts, success becomes more unlikely. We leave a thorough characterization of these properties to future work and discuss empirically in Appendix \ref{sec:privacy} the danger of leaking the marginal label distribution in \fedmix{} and \fedavg{}.

\begin{algorithm}[htb]
\caption{The \fedmix{} algorithm. $\alpha, \beta$ are the client and server learning rates; $\gamma$ is a dampening factor and $Z$ a normalization constant.}\label{alg:fedmix}
\begin{algorithmic}
\Function{Server side}{}
    \State Initialize $\bs{\phi}$ and $K$ vectors $\bv{W} = [\bv{w}_1, \dots, \bv{w}_K]$ 
    \For{round $t$ in $1,\dots T$}
        \State $S' \gets \text{random subset of the clients}$
        \State Initialize $\Delta_{\bv{W}}^t = \mathbf{0}, \Delta_{\bs{\phi}}^t = \mathbf{0}$
        \For{$s$ in $S'$}
            \State $\bv{W}^t_{s}, \bs{\phi}^t_s, p(z|s)\gets$ \Call{Client side}{$s, \bs{\phi}, \bv{W}$}
        \EndFor
        \State $p(s|z) \gets p(z|s) p(s)/\sum_{s\in S'} p(z|s)p(s) $
        \For {$s$ in $S'$}
            \State $\Delta_{\bv{w}_{k}}^t += p(s|z=k) (\bv{w}^{t-1}_{k} - \bv{w}^t_{s, k}) \enskip \forall k$
            \State $\Delta_{\bs{\phi}}^t += \frac{N_s}{N_{S'}}(\bs{\phi}^{t-1} - \bs{\phi}^t_{s})$
        \EndFor
        \State $\Delta_{\bs{\phi}}^t -= \nabla_{\bs{\phi}}H(\sum_c\!q_\phi(z|y\!=\!c)p(y\!=\!c))$
        \State $\bv{w}^{t+1}_{1:K} \gets$ \Call{Adam}{$\Delta_{\bv{w}_{1:K}}^t, \beta$}
        \State $\bs{\phi}^{t+1} \gets$ \Call{Adam}{$\Delta_{\bs{\phi}}^t, \beta$}
    \EndFor
\EndFunction
\vspace{0.1cm}
\Function{Client side}{$s, \bs{\phi}, \bv{W}$}
    \State Get local parameters $\theta_s$
    \For{epoch $e$ in $1, \dots, E$}
        \For {batch $b \in B$}
            \State $\phi_c' = p_{\bv{w}_z,\theta_s}(y_b=c,z|\bv{x}_b,s)^{1/(\beta N_{s,c})}/Z$
            \State $\phi_c \gets \gamma \phi_c + (1-\gamma)\phi_c' $ \Comment{Dampening}
            \State $L_s \gets \mathbb{E}_{q_\phi(z|y_b)}[\log  p_{\bv{w}_z,\theta_s}(y_b,z|\bv{x}_b,s)]$
            \State $\bv{W} += \alpha \nabla_{\bv{W}}L_s$
            \State $\theta_s += \alpha \nabla_{\theta_s} L_s$
        \EndFor
    \EndFor
    \State $q(z|s) \gets \mathbb{E}_{y\sim \mathcal{D}_s}[q_\phi(z|y)]$\\
    \enskip\enskip\enskip\Return $\bv{w}_{1:K}, \bs{\phi}, q(z|s)$
\EndFunction

\end{algorithmic}
\end{algorithm}
\paragraph{Communication costs}
Reducing the amount of communication via the internet plays a central role, especially in the cross-device setting of FL. \fedmix{} directly increases communication by a factor of $K$ and is therefore more suited to the cross-silo setting of FL. Nevertheless, \fedmix{} offers several possible avenues for reducing communication costs. One such way is to ``prune away'' experts locally from the MoE if $q_\phi(z|s)$ does not surpass a threshold $\eta/K$. Alternatively, since each expert is only modelling a subset of the data-space, their required modelling capacity and therefore parameter size can be reduced compared to the \fedavg{} model. Finally, sharing those parameters between experts that exhibit small gradient divergence or via a parameter efficient construction, such as with rank-1 factors \cite{dusenberry2020efficient}, are viable options to reduce communication. We explore pruning of experts in Appendix \ref{sec:CommCosts} and leave alternatives to future work. 

\paragraph{Designing robust gates}
In the federated scenario, $N_s$ is often much smaller than $N$ and especially small in relation to the complexity of the data we try to model. Any localized parameters therefore are prone to overfitting. On the other hand, the global parameters of an expert are trained using all data-points assigned to that expert across all shards, allowing to learn more robust features. 

We can make use of the robustness of these experts' features for the gating mechanism by conditioning on them instead of training an entirely separate model for $p_{\theta_s}(z| \bv{x}, s)$. Let us define $h_k(\bv{x})$ as intermediary features of expert $k$. In order to scale with the number of experts, we introduce the local vector $\bm{\pi}_s\in \mathbb{R}_+^K, \sum_k^K\pi_{k,s}=1$ with which the intermediate features are averaged before applying a linear transformation to compute the input to the softmax gates: 
\begin{align}\label{eq:gate_design}
    h_s(\bv{x}) &= \sum_{k=1}^K \pi_{k,s}h_k(\bv{x}) \nonumber \\ 
    p_{\theta_s}(z|\bv{x},s) &= \text{SM}\left(\bv{A}_s^Th_s(\bv{x}) \!+\! \bv{b}_s\right)
\end{align}
where $\theta_s = (\bm{\pi}_s,\bv{A}_s, \bv{b}_s)$ are local learnable parameters and $\text{SM}$ represents the softmax function.

\paragraph{Inference at test time}
We consider three variants for test-time evaluation of \fedmix{}. In the first case, a client $s$ that participated in training is presented with a new data point $(\bv{x}^*,s)$. Predictions can then be straightforwardly done by selecting the $y$ that maximizes $\sum_{z=1}^K p(y|\bv{x}^*,z)p(z|\bv{x}^*,s)$.
In the second, more challenging, scenario a new client $s^*$ is introduced together with a new labelled local data set $\mathcal{D}_{s^*}$. Here we propose to instantiate and train the local gating mechanism by optimizing the parameters $\theta_s$ of $p_{\theta_s}(z|\bv{x},s^*)$ via the local objective. Afterwards, predictions can be made in a manner similar to the first case.

Finally, we consider the case in which a new client $s^*$ has no labelled dataset available. Without a local gating function, simply ensembling experts exhibits almost random behaviour since experts can be overly confident on out-of-distribution data~\citep{snoek2019can}. We therefore propose to ensemble across local gating mechanisms to compute $p(z|\bv{x}^*) = \sum_{s=1}^S p_{\theta_s}(z|\bv{x}^*,s)p(s)$; a method which works well in practice. In Appendix \ref{sec:NewShardInf} we discuss a more principled approach based on a complete graphical model perspective.

\section{Related Work}
\fedmix{} has similarities to many recent works in the topic of federated learning. Two methods closely related to ours are described in \citep{sattler2019clustered,briggs2020federated}. The authors propose to perform hierarchical clustering on the updates returned from each shard in order to incrementally create separate models for groups of users, with a cluster assignment mechanism based on handcrafted heuristics. \fedmix{} instead takes a different approach; it starts with a fixed set of $K$ models and then optimizes with gradient descent at each shard a per-datapoint model assignment mechanism that can better fit the peculiarities of the local dataset. 

Another closely related work is presented by \cite{mansour2020three}, where the authors propose to similarly create an ensemble of $K$-models and assign to each shard the model that achieves the lowest training loss on the local dataset. This is closer to the assignment that happens in \fedmix{} with one main difference; \fedmix{} takes into account the uncertainty in the selection mechanism as well with $p(z|\bv{x} ,s)$ instead of selecting the top performing component during training. This is beneficial early in training where the models have not fully specialized yet. Using local and global model parts has also been explored by~\cite{liang2020think}. The authors propose to have a local feature extractor at each shard and a global classifier on top of those features as opposed to having $K$-separate models and a local selection mechanism as in \fedmix{}. This setup yields improvements upon the vanilla federated averaging algorithm, however there are two potential drawbacks; first, empirically, the authors had to start their procedure from a pre-trained model with \fedavg{} and secondly, they have to ensemble all of the different feature extractors for predictions in new shards. In our experiments, we omit the pre-training step and show that the ensembling strategy fails. 

Federated learning in the non-i.i.d setting can also be improved upon in other ways. \cite{li2018federated} propose to employ a proximal regularizer at the shard level in order to prevent the local models from drifting too far from the global model, thus making federated learning more robust. \cite{jiang2019improving} notice that \fedavg{} and Reptile~\citep{nichol2018first}, a meta-learning algorithm, are essentially the same algorithm and thus propose fine tuning with Reptile in order to improve the personalized performance of the global model. In a similar vein, there are promising new works that explore the meta-learning view of federated learning~\citep{chen2018federated,khodak2019adaptive,fallah2020personalized}. Improving the personalized performance of the global model has also been done without meta-learning in works such as by ~\cite{deng2020adaptive,mansour2020three}.
In general, such improvements are complementary to \fedmix{} and can be used to further enhance its performance. We refer the interested readers to the recent surveys by~\citep{kairouz2019advances,kulkarni2020survey}.

\begin{table*}[ht]
\setlength{\tabcolsep}{3.0pt}
\centering
\caption{Average test-set accuracies across clients and communication costs (rounds and GB) for Cifar10, Cifar100 and Femnist at the end of training. We report both, the local accuracy after fine-tuning for one local epoch as well as the accuracy when evaluating at the server across the union of all local test-sets.}
\small
\vspace{0.12in}
\begin{tabular}{l|ccc|ccc|ccc}
\toprule
Method & \multicolumn{3}{c|}{Cifar 10} & \multicolumn{3}{c|}{Cifar 100} &  \multicolumn{3}{c}{Femnist} \\
& Local & Global & Comm. & Local & Global & Comm. & Local & Global & Comm.\\
\midrule
\fedavg{} & $85.98\%$ & $69.91\%$ & $2k, 137.3GB$ & $65.67\%$ & $46.97\%$ & $10k, 205.53GB$ & $90.93\%$ & $86.26\%$ & $5.2k, 235.71GB$\\
\textit{biased} \fedavg{} & $86.82\%$ & $70.62\%$ & $2k, 130.92GB$ &$64.41\%$ & $43.33\%$ & $10k, 205.52GB$ & $90.72\%$& $85.83\%$ & $5.2k, 235.71GB$ \\
Local/Global & $83.13\%$ & $10.00\%$ &$2k, 129.93GB$ &$42.82\%$& $6.60\%$& $6k, 116.73GB$ & $85.54\%$ & $2.17\%$ & $5.2k, 235.61GB$ \\
\fedmix{} K=4 & $\mathbf{88.54\%}$ & $\mathbf{76.42\%}$ & $2k, 522.5 GB$ & $\mathbf{67.54\%}$ & $\mathbf{47.48\%}$  & $10k, 822.12GB$ & $\mathbf{91.47\%}$& $\mathbf{87.00\%}$ & $5.2k, 942.84GB$\\
\bottomrule
\end{tabular}
\label{tab:acc_comm_res}
\end{table*}

\section{Experiments}
\label{sec:experiments}
We evaluate \fedmix{} with $K=4$ experts across several datasets and non-i.i.d. settings. Results with different number of experts $K$ can be found in Appendix \ref{sec:ablation}. For label-skew we use Cifar10, Cifar100 \citep{krizhevsky2009learning} and femnist \citep{caldas2018leaf}, a 62-way image classification problem on hand-written digits and letters that is also naturally non-i.i.d due to the different writing styles of 3500 users. In Appendix \ref{sec:ExpSetup} we detail the experimental setup and provide additional ablation studies in Appendix \ref{sec:ablation}. For non-i.i.d.-ness in the input, we perform experiments on rotated MNIST \citep{lecun2010mnist}. For differences in the classification mechanism $p(y|x,s)$ we explore label-permutation on Cifar10, following \cite{sattler2019clustered}, and show the strength of \fedmix{} as a federated clustering algorithm.

\begin{figure}[h]
\centering
 
  \includegraphics[width=\linewidth]{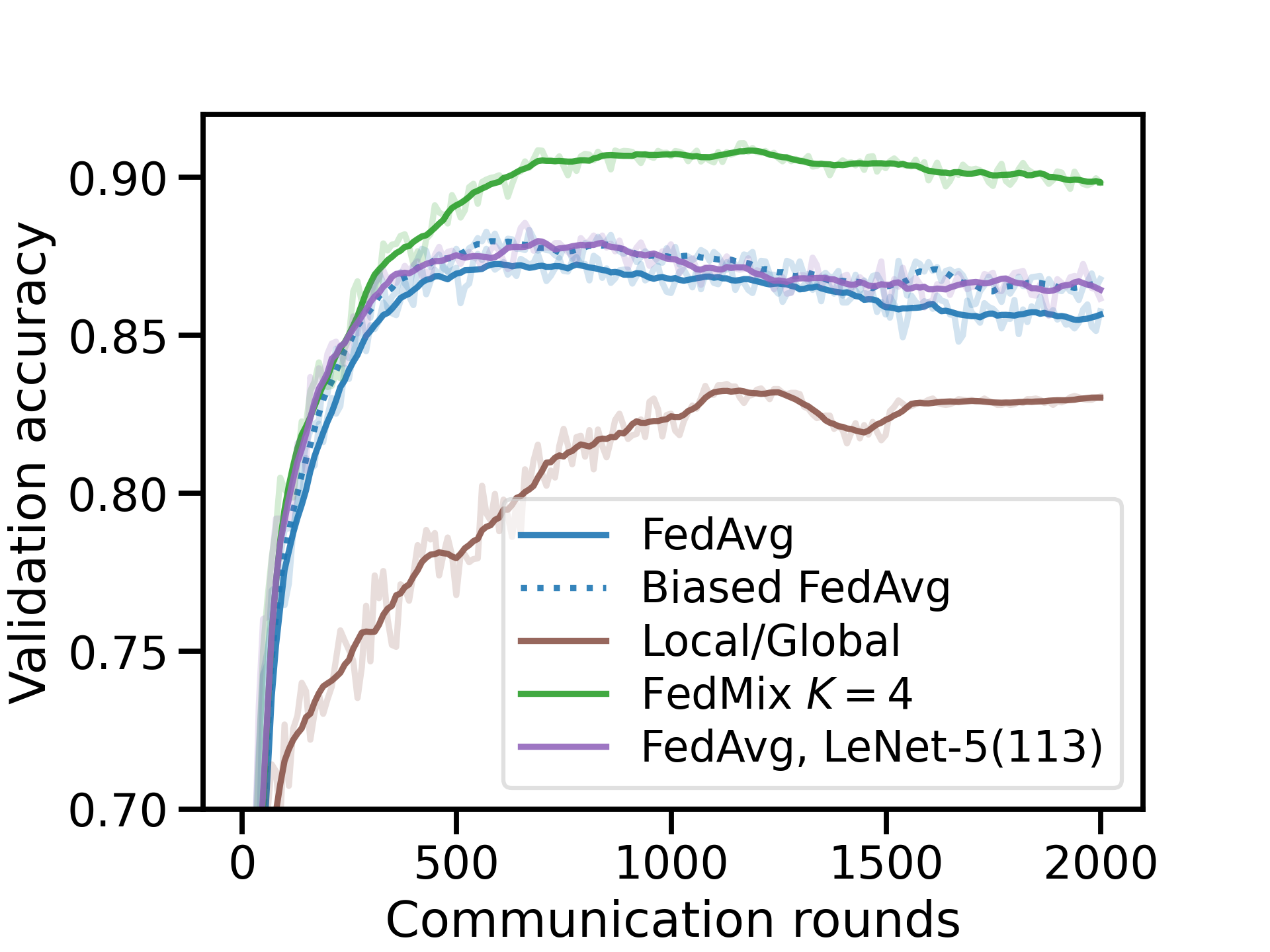}
  \caption{Average accuracy across all clients (y-axis) as a function of communication rounds for Cifar 10. Best viewed in color.}
  \label{fig:cifar10}
\end{figure}

\begin{figure}[h]
\centering
  \includegraphics[width=\linewidth]{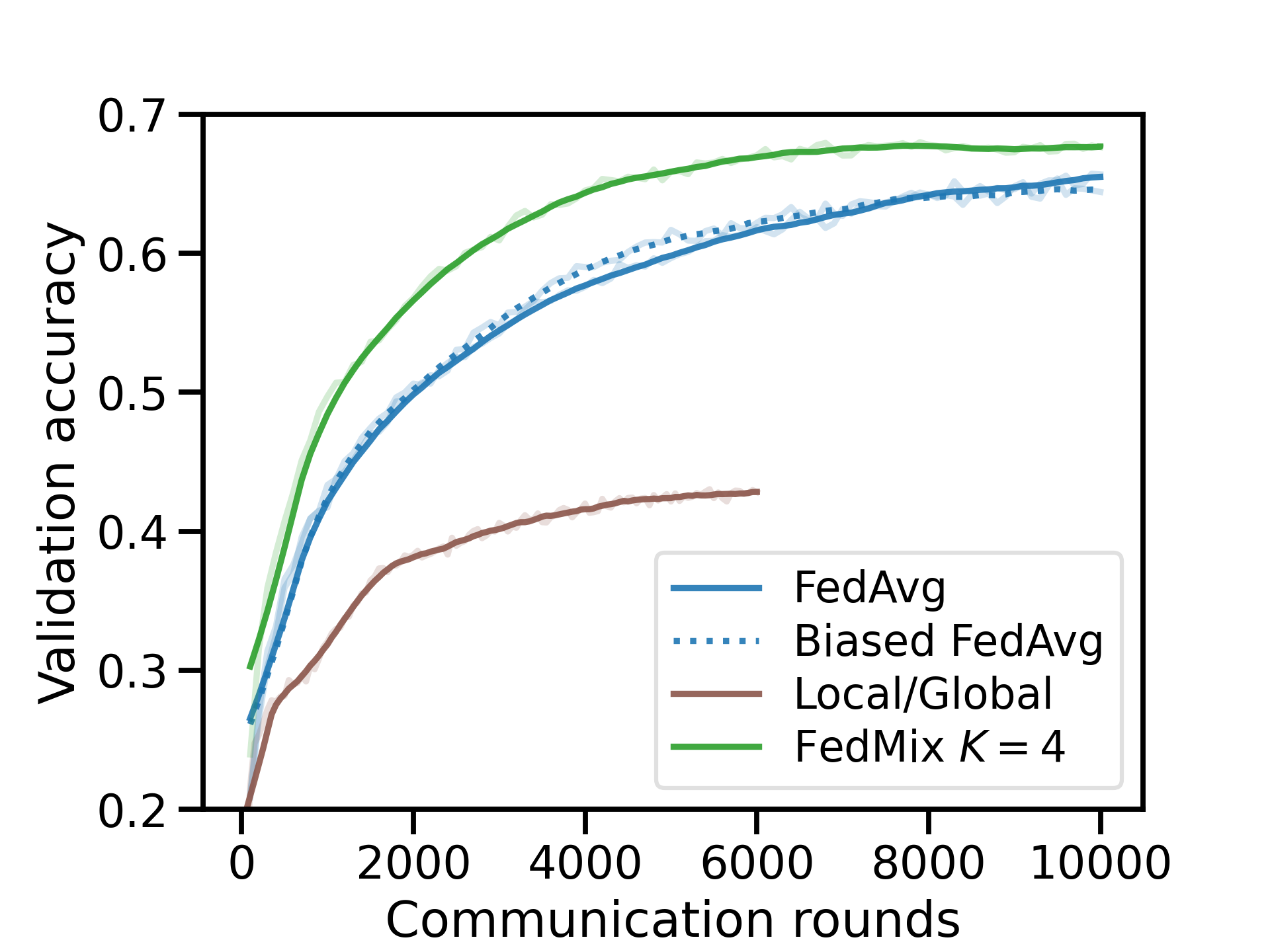}
  \caption{Average accuracy across all clients (y-axis) as a function of communication rounds for Cifar 100. Best viewed in color.}
  \label{fig:cifar100}
 \end{figure}

\begin{figure}[h]
\centering
  \includegraphics[width=\linewidth]{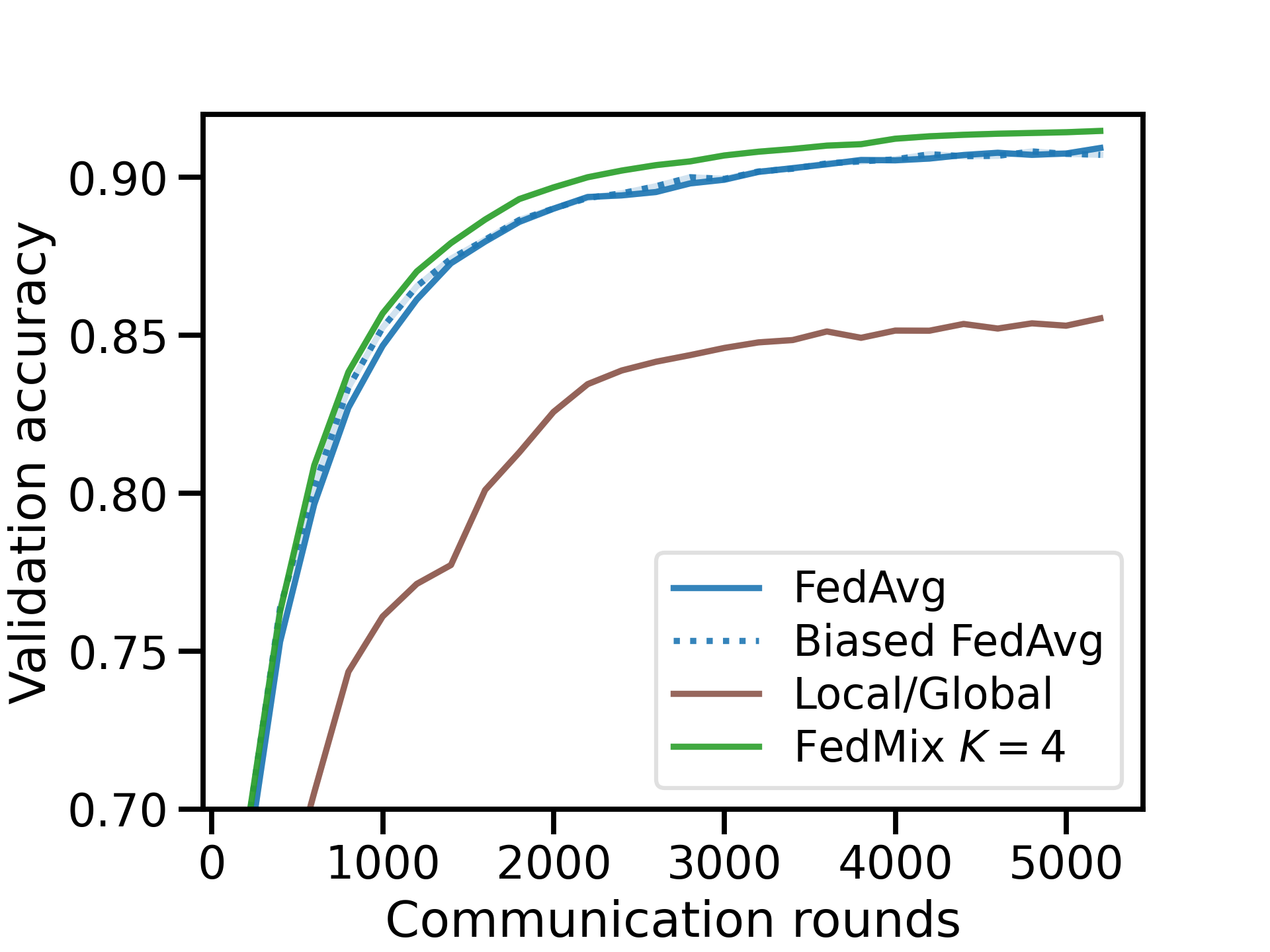}
  \caption{Average accuracy across all clients (y-axis) as a function of communication rounds for Femnist. Best viewed in color.}
  \label{fig:femnist}
\end{figure}
\subsection{Label Skew}
We compare \fedmix{} along several dimensions to baselines such as (generalized) \fedavg{} \citep{reddi2020adaptive}, \textit{biased} \fedavg{}, and the Local/Global approach of \cite{liang2020think}.  In \textit{biased} \fedavg{}, we allow each client to learn a personalized bias vector $b_s$ of its output layer. This will allow \textit{biased} \fedavg{} to model the label skew at each client but, fundamentally, cannot model any other form of non-i.i.d.-ness. Similarly, \cite{liang2020think} propose to split the model into local and global components by having local feature extractors and learning the upper layers of the neural network via \fedavg{}. We experimented with splitting LeNet-5 at every intermediate layer and report results with the best performing split: keeping the input layer local. For ResNet-20, splitting after the first block performed best. We show that training with \fedmix{} achieves higher personalized model accuracy as well as global model accuracy on the server. 

Figures \ref{fig:cifar10},~\ref{fig:cifar100} and~\ref{fig:femnist} shows learning curves for these different settings. These curves were obtained by averaging the local clients' model accuracy on the validation set using the model that they last communicated to the server, thereby serving as a proxy for the local models' performance. 
Table \ref{tab:acc_comm_res} shows that \fedmix{} consistently outperforms the other approaches at the cost of more GBs communicated. Figure \ref{fig:cifar10} shows, however, that 
for the same communication costs per communication round, \fedmix{} offers better performance than \fedavg{}, as can be seen in comparing \fedmix{} to \fedavg{} LeNet-5 with 113 channels. The channel count has been chosen such that the model size approximates $4$ separate standard LeNet experts respectively. 
The advantage of \textit{biased} \fedavg{} and Local/Global shows if we do not allow a local client to fine-tune the global model parameters. This setting becomes relevant if a clients' local training data does not generalize to the test data, either because of a distribution shift of because it is too small. With finetuning we find that \fedmix{} outperforms these approaches given that the local training data is sufficient. We show learning curves with the server-models in Appendix \ref{sec:server_w}.

For Femnist we observe a less dramatic improvement in performance compared to \fedavg{} since its source of non-i.i.d.-ness is not only expressed through label-skew.

\subsection{Rotated MNIST}
\label{sec:rotMnist}
\begin{figure}[h]
\centering
        \includegraphics[width=\linewidth]{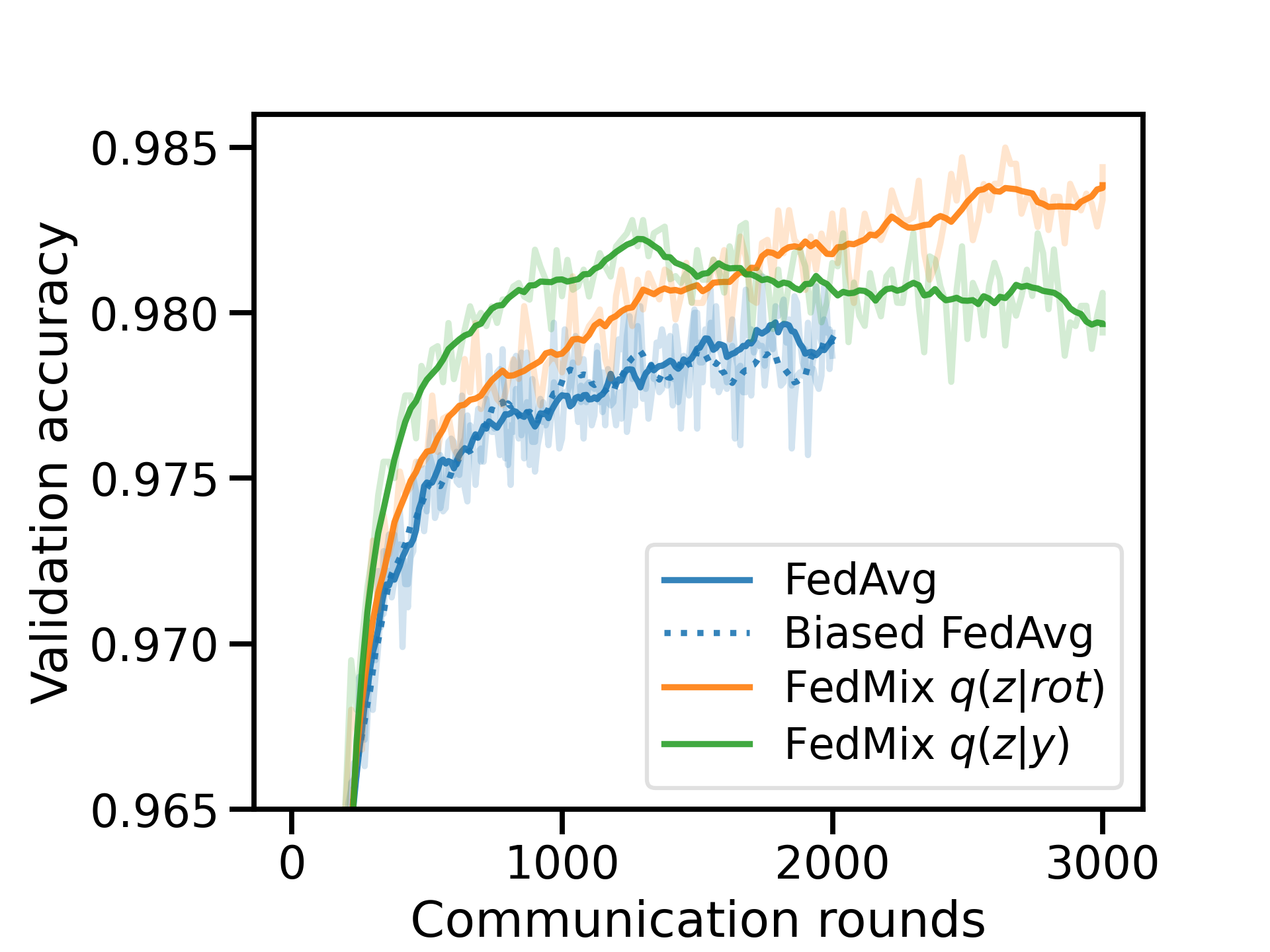}
        \caption{Only rotation}
        \label{fig:rotMnistAlphaInf}
\end{figure}

\begin{figure}[h]
\centering
        \includegraphics[width=\linewidth]{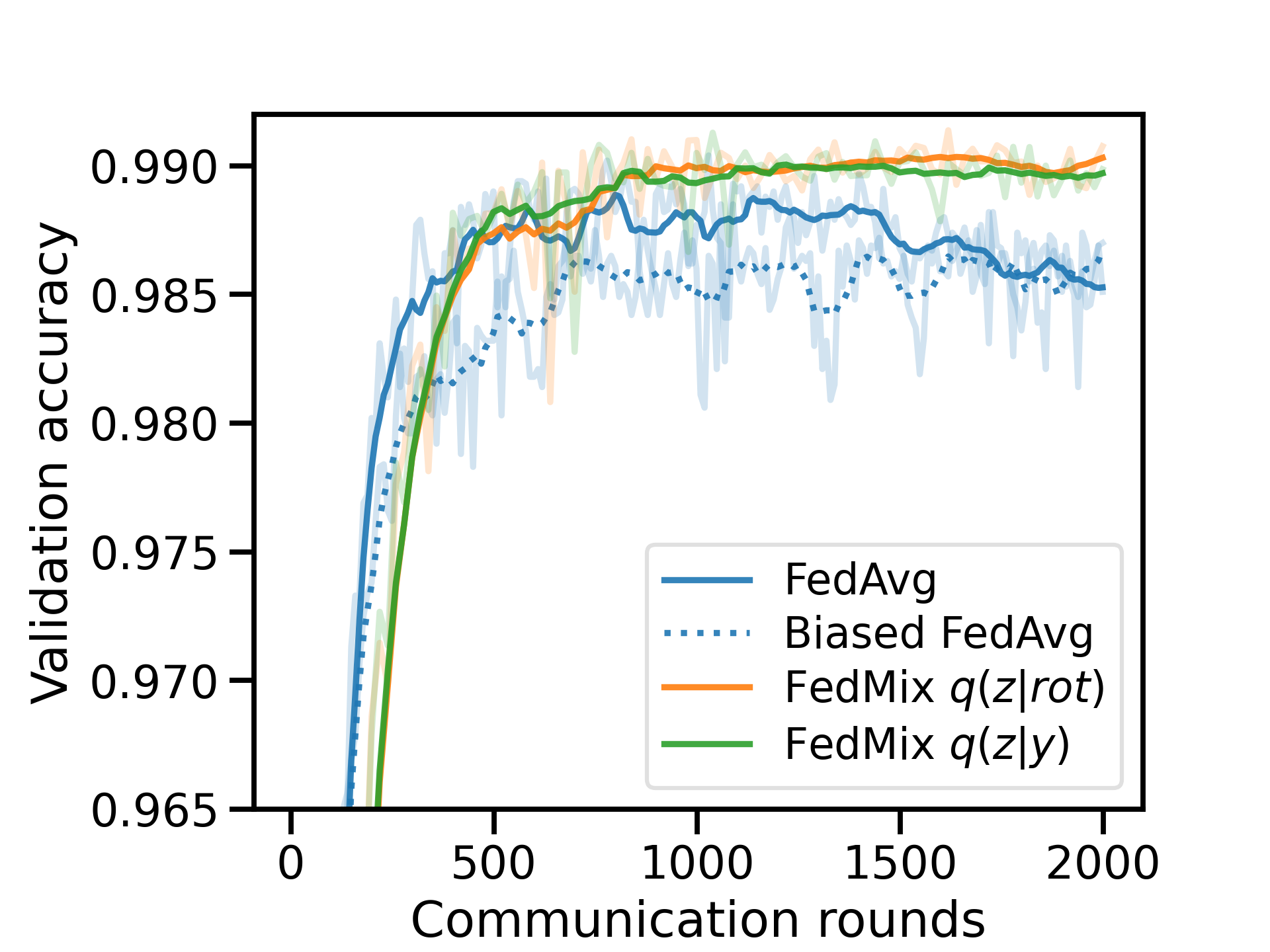}
        \caption{Rotation and labels}
        \label{fig:rotMnistAlpha1}
\end{figure}

To show that \fedmix{} is not limited to label-skew, we create a federated rotated MNIST dataset with 100 clients. Instead of label skew, each client randomly chooses a multiple of $45$ degrees from a different probability distribution over $8$ possible rotation angles to rotate a digit with. Each client's distribution is drawn from Dir($\alpha=1.0$). At test time, each data-point is randomly rotated according to the client's distribution. Additionally, we create a dataset where instead of uniform sampling of labels, we replicate the non-i.i.d. label skew described for Cifar10 above and combine it with the rotation non-i.i.d.-ness. 

We compare \fedmix{} where $q$ is conditioned on $y$ or on the degrees of rotation for a data-point against baselines. Figure \ref{fig:rotMnistAlphaInf} shows that \fedmix{} benefits from being conditioned on the correct side-information for this task. Conditioning on $y$ initially improves performance, however degrades in the end. In the presence of both sources of non-i.i.d.-ness, Figure \ref{fig:rotMnistAlpha1} shows that \fedmix{} improves performance regardless of which conditioning is chosen.

\begin{figure}[H]
    \centering
    \includegraphics[trim=0cm 0cm 0cm 1cm,width=0.9\linewidth]{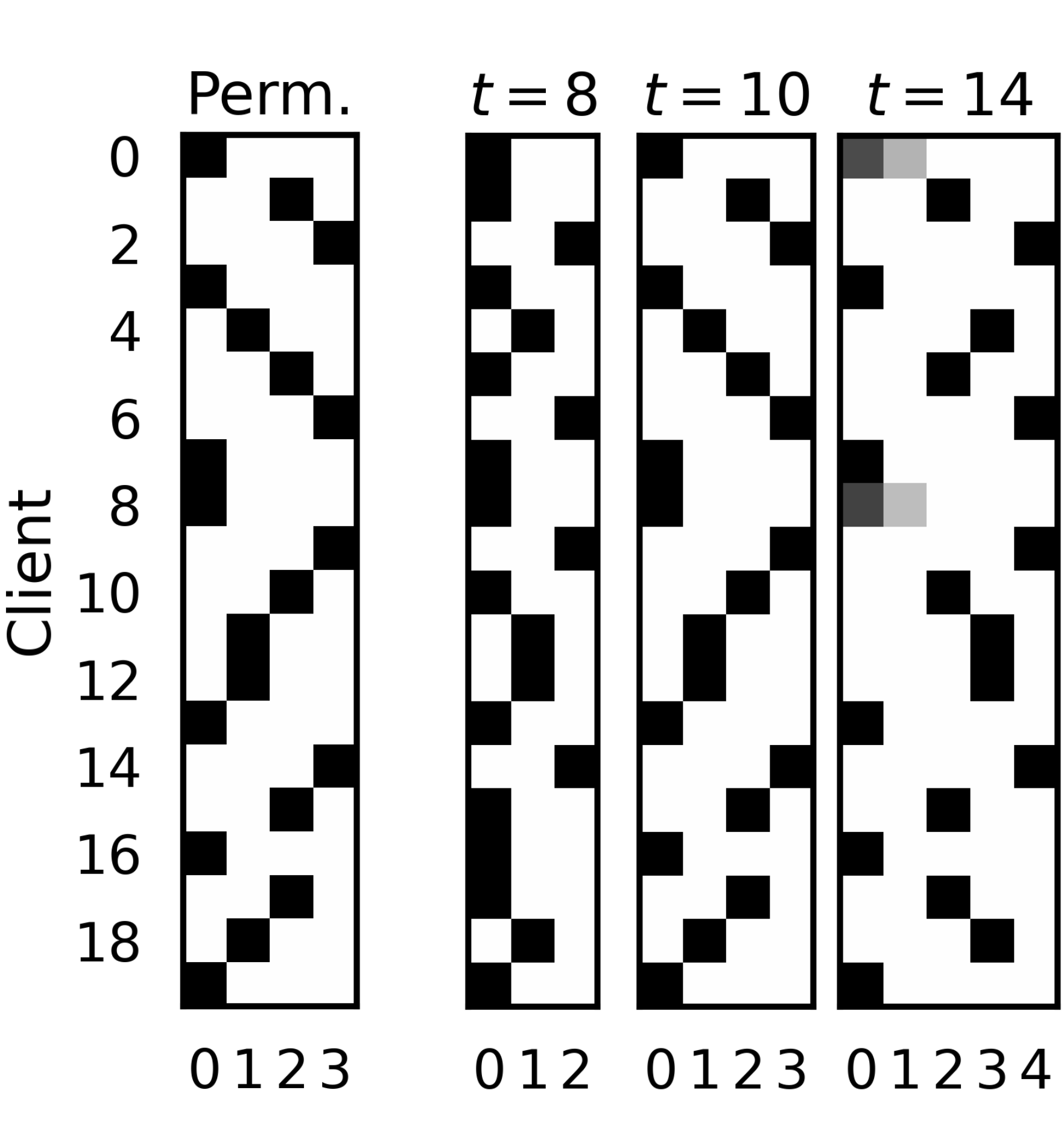}
    \caption{Ground truth and $q(z|s)$ for $K \in \{3,4,5\}$ after different number of communication rounds $t$. The ordering of columns is arbitrary. Greyscale represents probabilities (white: $0$; black: $1$)}
    \label{fig:perm_development}
\end{figure}
\subsection{Label Permutations}
\label{sec:expPerm}
Apart from non-i.i.d.-ness in $p(y)$ and $p(x)$, we can expect the mapping $p(y|x)$ itself to be different between clients. We replicate the experimental setup of \citep{sattler2019clustered} with 20 clients for Cifar10, $C=1.0$ and $E=3$, albeit with LeNet-5. Each client is randomly assigned one of $4$ different label permutations, determining the cluster assignment $q(z|s)$ that \fedmix{} has to learn. Although \fedmix{} is designed to distinguish different regions of the input space, we show that it can perform user clustering. The gating function $p(z|x,s)$ learns to correctly identify, for each data point, the expert corresponding to the permutation of $s$, thus recovering the original clustering; Figure \ref{fig:perm_development} illustrates this effect (please note that the ordering of columns is arbitrary) for different number of experts $K$, using a dampening value of $\gamma=0.75$. 

When $K$ matches the number of clusters in the data generating process, we see that \fedmix{} correctly identifies the user clustering after just $10$ communication rounds, which is much faster than the results presented in~\cite{sattler2019clustered}. In the cases where $K$ differs from the ground truth number of clusters we observe two phenomena, depending on whether we have fewer or more experts. When $K$ is smaller, \textit{i.e.}, $K=3$, we observe that after $8$ rounds, one expert takes responsibility for two of the four permutations whereas the other two experts correctly identify the remaining clusters. In the case of more experts than clusters, \textit{i.e.}, $K=5$, we observe that after $14$ rounds \fedmix{} splits the responsibility of one permutation across two experts and uses the other three experts to model each of the remaining permutations. Overall, we observe that \fedmix{} has intuitive behaviour and, given enough capacity, can correctly identify the label permutations of each client.

\section{Discussion}
With \fedmix{} we have introduced a federated learning algorithm that explicitly takes the non-i.i.d. characteristics of a federated dataset into account. We showed \fedmix{}' strength across a variety of non-i.i.d. datasets ranging from label-skew to input transformations and label permutations. Clients can learn to align specialized experts on sub-regions of the data space and achieve higher performance compared to \fedavg{}, in situations where the source of the non-i.i.d. nature is known. This assumption is very strong in real-world federated scenarios and we expect a more flexible alignment process than a global $q$ to be the most interesting avenue for future research. 
In the future, we will explore ways to perform automatic selection of $K$ as well as automatic selection of architecture elements to share between experts, trading-off gradient alignment and communication budgets.

\bibliography{bibliography}
\bibliographystyle{bibliography}

\newpage
\onecolumn
\appendix
\newpage
\onecolumn
\appendix

\section{Experimental Setup}
\label{sec:ExpSetup}
For all experiments in the experimental section, we use a SGD optimizer with a learning rate of $0.05$ locally and the Adam~\citep{kingma2014adam} optimizer with its default hyperparameters at the server by interpreting the difference of the local from the global model as a gradient~\citep{reddi2020adaptive}. For \fedmix{}, the features  $h_k(\bv{x})$ are defined as the input to a expert $k$'s output layer. Unless otherwise mentioned, we choose $\gamma=0.99$ for the dampening value of the $\phi_{c,s}$ updates locally.

\subsection{Label Skew}
\paragraph{Cifar10}
For Cifar10, we replicate the federated data split of \cite{hsu2019measuring}. The dataset is split across $100$ clients, whose data-points are drawn according to their label from a $\text{Dir}(\alpha = 1.0)$ distribution without replacement. For the base model, we use a LeNet-5 architecture \citep{lecun1998gradient}. We use a batch size of $B = 64$ locally. We sample $10$ clients without replacement on each round (but with replacement across rounds) and train for $E = 1$ local epochs. We found dropout to be necessary to avoid overfitting to the local training data-sets. For all experiments with the LeNet-5 architecture on Cifar10 we found a dropout rate of 0.3 for the second convolutional layer and a rate of 0.1 for the first fully connected layer to be helpful, except for the Local/Global \cite{liang2020think} experiments, where we consequently omit dropout. We found a value of $\beta=0.8$ to be best in balancing between robust experts and fast specialization. 

\paragraph{Cifar100}
For Cifar100 we replicate the data split of~\cite{reddi2020adaptive}. The dataset is split into $500$ clients by using a hierarchical model over the coarse and fine labels, with the same hyperparameters as the ones provided by~\cite{reddi2020adaptive}. The other hyperparameters are the same as Cifar10 with the exception of the batch size, where we use $B=20$, as well as the architecture, where we use a ResNet-20 with group normalization~\citep{wu2018group} layers instead of batch normalization~\citep{ioffe2015batch}. We augment the data by random cropping from a $4$ pixel padded image and horizontal flipping. We found $\beta=1.0$ to perform well for Cifar100, resulting in reliable specialisation. 
\paragraph{Femnist}
Finally, for the Femnist dataset, we similarly followed the setup of~\cite{reddi2020adaptive} with the same LeNet-5 architecture and hyperparameters of Cifar10 with the exception of the batch size where we used $B=20$. We found no danger of overfitting, so we omit dropout here. Similarly to Cifar10, we found a value of $\beta=0.8$ to serve as a good tradeoff.

\subsection{Rotated Mnist}
For the experiments on rotated Mnist, we use the LeNet-5 architecture \citep{lecun1998gradient} and keep the same hyperparameters as for Cifar10, \textit{i.e.} a local batch size of $B=64$, $10$ clients per round and a single local epoch $E=1$ but do not use dropout. When conditioning on the label information \textit{i.e.} using $q(z|y)$, we choose $\beta=0.8$ and $\gamma=0.99$. When using $q(z|rot)$, we found such a high dampening factor does not lead to any specialization. Instead, lowering it to $\gamma=0.5$ achieved specialized experts.

\subsection{Label Permutations}
For the label permutation experiments, we again make use of LeNet-5 on the Cifar10 dataset. Replicating the setup of \citep{sattler2019clustered}, we deviate from our previous hyperparameters and chose $S=20$ clients, no client subsampling (\textit{i.e.} $C=1.0$) and three local epochs $E=3$ per client. We omit dropout since we do not train these models to convergence. We found \fedmix{} to be very robust in finding the correct cluster assignment in a small amount of steps. A value of $\gamma=0.75$ quickly causes specialization across all values of $K$ we experiment with.

\section{Solving for $\phi_c$}
\label{sec:fixedformmath}
Here we present the derivation for $\phi$ discussed in the main text. We aim to maximize Eq. \ref{eq:LowerBound} on a specific client $s$ for the parameters $\phi$ of $q_\phi(z|y)$, \textit{i.e.} solve $\argmax_\phi L_s(\phi)$ with
\begin{align}
    L_s(\phi) = \sum_{i=1}^{N_s}\mathbb{E}_{q_\phi(z|y_{s,i})}[
\log p_{\bv{w}_z}(y_{s,i}|\bv{x}_{s,i},z)p_{\theta_s}(z|\bv{x}_{s,i},s)]+\beta H(q_\phi(z|y_{s,i})).
\end{align}
We assume a parameterization in the form of $q(z=k|y=c) = \phi_{c,k}, \sum_{z=1}^K \phi_{c,k} = 1$. We can extend out maximization target to 
\begin{align}
    L_s(\phi) &= \sum_{c=1}^C \sum_{i=1}^{N_{s,c}}\mathbb{E}_{q_\phi(z|y_{s,i}=c)}[
\log p_{\bv{w}_z}(y_{s,i}=c|\bv{x}_{s,i},z)p_{\theta_s}(z|\bv{x}_{s,i},s)]+\beta H(q_\phi(z|y_{s,i}))\\
    &= \sum_{c=1}^C \sum_{i=1}^{N_{s,c}}\left( \sum_{k=1}^K \phi_{c,k}[
\log p_{\bv{w}_z}(y_{s,i}=c|\bv{x}_{s,i},z)p_{\theta_s}(z|\bv{x}_{s,i},s)]-\beta \sum_{k=1}^K \phi_{c,k}\log \phi_{c,k}\right),
\end{align}
where $C$ corresponds to the number of classes, $N_{s,c}$ the number of data-points on shard $s$ belonging to class $c$ and in the second line we have substituted $q$ for its direct parameterization. 

The requirement of $\sum_{z=1}^K \phi_{c,k} = 1, \forall c$ transforms the optimization problem into a constrained optimization problem that we can approach by introducing lagrangian multipliers $\lambda_c$ and optimizing the Lagrangian, \textit{i.e.} $\argmax_{\phi,\lambda} \tilde{L}_s(\phi,\lambda)$:
\begin{align}
    \tilde{L}_s(\phi,\lambda) = \sum_{c=1}^C \sum_{i=1}^{N_{s,c}}\left( \sum_{k=1}^K \phi_{c,k}[\log p_{\bv{w}_z}(y_{s,i}=c|\bv{x}_{s,i},z)p_{\theta_s}(z|\bv{x}_{s,i},s)]-\beta \sum_{k=1}^K \phi_{c,k}\log \phi_{c,k}\right) + \sum_{c=1}^C \lambda_c \sum_{k=1}^K (\phi_{c,k}-1).
\end{align}

The problem can be decomposed into $C$ independent problems $\argmax_{\phi_c,\lambda_c}\tilde{L}_{s,c}(\phi_c,\lambda_c)$, solving for $\phi_c$ and $\lambda_c$ each. Setting the gradient with respect to the lagragian multiplier to zero recovers our constraint:
\begin{align}
    \frac{\partial}{\partial\lambda_c} \tilde{L}_{s,c}(\phi_c,\lambda_c) & = 0\\
    \sum_{z=1}^K \phi_{c,k} &= 1 \label{eq:sum_constraint}
\end{align}

We compute the gradient $\frac{\partial}{\partial\phi_{c,k}}\tilde{L}_{s,c}(\phi_c,\lambda_c)$ as follows, set it to zero and solve for $\phi_{c,k}$:
\begin{align}
 \frac{\partial}{\partial\phi_{k,c}}\tilde{L}_{s,c}(\phi_c,\lambda_c) &=\sum_{i=1}^{N_{s,c}}\left[\log p_{\bv{w}_z}(y_{s,i}=c|\bv{x}_{s,i},z)p_{\theta_s}(z|\bv{x}_{s,i},s) - \beta (\log \phi_{c,k} + 1)\right] + \lambda_c = 0\\
 &=\sum_{i=1}^{N_{s,c}}\left[\log p_{\bv{w}_z}(y_{s,i}=c|\bv{x}_{s,i},z)p_{\theta_s}(z|\bv{x}_{s,i},s)\right] - \beta N_{s,c} \log \phi_{c,k} - \beta N_{s,c} + \lambda_c = 0\\
 \log \phi_{c,k} &= \frac{1}{\beta N_{s,c}}\left(\sum_{i=1}^{N_{s,c}}\left[\log p_{\bv{w}_z}(y_{s,i}=c|\bv{x}_{s,i},z)p_{\theta_s}(z|\bv{x}_{s,i},s)\right] -\beta N_{s,c} + \lambda_c\right)\\
 \phi_{c,k} &= \exp{\left(\frac{\lambda_c}{\beta N_{s,c}} - 1\right)} \left(\prod_{i=1}^{N_{s,c}} p_{\bv{w}_z}(y_{s,i}=c|\bv{x}_{s,i},z)p_{\theta_s}(z|\bv{x}_{s,i},s)\right)^{\frac{1}{\beta N_{s,c}}} \label{eq:solve_for_phi}
\end{align}

Combining the last result for $\phi_{c,k}$ with Eq. \ref{eq:sum_constraint}, we can proceed to solve for $\lambda_c$:
\begin{align}
    \sum_{z=1}^K \phi_{c,k} = 1&= \exp{\left(\frac{\lambda_c}{\beta N_{s,c}} - 1\right)} \sum_{z=1}^K \left(\prod_{i=1}^{N_{s,c}} p_{\bv{w}_z}(y_{s,i}=c|\bv{x}_{s,i},z)p_{\theta_s}(z|\bv{x}_{s,i},s)\right)^{\frac{1}{\beta N_{s,c}}}  \\
    \exp{\left(1-\frac{\lambda_c}{\beta N_{s,c}}\right)}&=\sum_{z=1}^K \left(\prod_{i=1}^{N_{s,c}} p_{\bv{w}_z}(y_{s,i}=c|\bv{x}_{s,i},z)p_{\theta_s}(z|\bv{x}_{s,i},s)\right)^{\frac{1}{\beta N_{s,c}}}\\
    1-\frac{\lambda_c}{\beta N_{s,c}} &= \log \left( \sum_{z=1}^K \left(\prod_{i=1}^{N_{s,c}}  p_{\bv{w}_z}(y_{s,i}=c|\bv{x}_{s,i},z)p_{\theta_s}(z|\bv{x}_{s,i},s)\right)^{\frac{1}{\beta N_{s,c}}} \right)\\
    \lambda_c &= \beta N_{s,c} \left(1- \log \left( \sum_{z=1}^K \left(\prod_{i=1}^{N_{s,c}}  p_{\bv{w}_z}(y_{s,i}=c|\bv{x}_{s,i},z)p_{\theta_s}(z|\bv{x}_{s,i},s)\right)^{\frac{1}{\beta N_{s,c}}} \right) \right)
\end{align}

Finally, we can integrate our solution for $\lambda_c$ into Eq. \ref{eq:solve_for_phi}. Note that the exponential expression containing $\lambda_c$ simplifies by cancelling out $\beta N_{s,c}$ and the factor $1$:
\begin{align}
    \phi_{c,k} &= 
    \exp{\left( - \log \left( \sum_{z=1}^K \left(\prod_{i=1}^{N_{s,c}}  p_{\bv{w}_z}(y_{s,i}=c|\bv{x}_{s,i},z)p_{\theta_s}(z|\bv{x}_{s,i},s)\right)^{\frac{1}{\beta N_{s,c}}} \right)\right)} \nonumber\\ &\quad\quad\quad\cdot \left(\prod_{i=1}^{N_{s,c}} p_{\bv{w}_z}(y_{s,i}=c|\bv{x}_{s,i},z)p_{\theta_s}(z|\bv{x}_{s,i},s)\right)^{\frac{1}{\beta N_{s,c}}}\\
&=\frac{\left(\prod_{i=1}^{N_{s,c}} p_{\bv{w}_z}(y_{s,i}=c|\bv{x}_{s,i},z)p_{\theta_s}(z|\bv{x}_{s,i},s)\right)^{\frac{1}{\beta N_{s,c}}}}{\left( \sum_{z=1}^K \left(\prod_{i=1}^{N_{s,c}}  p_{\bv{w}_z}(y_{s,i}=c|\bv{x}_{s,i},z)p_{\theta_s}(z|\bv{x}_{s,i},s)\right)^{\frac{1}{\beta N_{s,c}}} \right)}\\
&=\frac{\exp{\left( \frac{1}{\beta N_{s,c}} \sum_{i=1}^{N_{s,c}} \log p_{\bv{w}_z}(y_{s,i}=c|\bv{x}_{s,i},z)p_{\theta_s}(z|\bv{x}_{s,i},s) \right)}}{ \sum_{z=1}^K \exp{\left( \frac{1}{\beta N_{s,c}} \sum_{i=1}^{N_{s,c}} \log p_{\bv{w}_z}(y_{s,i}=c|\bv{x}_{s,i},z)p_{\theta_s}(z|\bv{x}_{s,i},s) \right)}} \label{eq:phi_solution},
\end{align}
where in the second step we have taken the exponent of the log of the product over $N_{s,c}$ data points, both, in the numerator and denominator. It is now easy to see how the solution for $\phi_{c,k}$ corresponds to a temperature $\beta$ modulated softmax over the log-likelihood of datapoints belonging to class $c$. 
In practice, a client $s$ performs mini-batch stochastic gradient descent on the parameters $\bv{w}_z,\theta_s$. We therefore want to avoid evaluating all $N_s$ data-points in order to compute $\phi_{c,k}$. Even though we only require the forward pass through our model, therefore not causing any memory issues, we want to avoid the additional computational overhead. Similarly to mini-batch stochastic gradient descent, we therefore approximate the costly average over $N_{s,c}$ datapoints in Eq. \ref{eq:phi_solution} by considering only a mini-batch of $M_s$ data-points, where $M_{s,c}$ denotes all data-points in the mini-batch $M_s$ that belong to class $c$:
\begin{align}
    \phi_{s,c}' = \frac{\exp{\left( \frac{1}{\beta M_{s,c}} \sum_{i=1}^{M_{s,c}} \log p_{\bv{w}_z}(y_{s,i}=c|\bv{x}_{s,i},z)p_{\theta_s}(z|\bv{x}_{s,i},s) \right)}}{ \sum_{z=1}^K \exp{\left(\frac{1}{\beta M_{s,c}} \sum_{i=1}^{M_{s,c}} \log p_{\bv{w}_z}(y_{s,i}=c|\bv{x}_{s,i},z)p_{\theta_s}(z|\bv{x}_{s,i},s) \right)}}.
\end{align}

In order to stabilize the update of $\phi_{s,c}$ from round $t$ to round $t+1$ during the local client's training epoch, we introduce a dampening factor $\gamma$,
\begin{align}
    \phi_{s,c}^{t+1} = \gamma \phi_{s,c}^{t} + (1-\gamma)\phi_{s,c}'^{t+1},
\end{align}
thereby completing the update rule for $\phi_{s,c}$.

\section{Privacy Implications}
\label{sec:privacy}
Privacy is one of the key motivations for research and deployment of Federated Learning. Even though privacy is not a focus of this paper, we briefly discuss some implications of making explicit use of $q(y|s)$ in \fedmix{} in the main text. Here, we want to shine light on the practical possibility to reconstruct $p(y|s)$ at the server-side in standard \fedavg{}, arguing that \fedmix{} does not reveal information that is not already accessible by the server.

Assume a randomly initialized model being sent to a client $s$, where the client performs a single full batch update step on the output layer's bias vector $b_s$. Assuming a softmax cross-entropy loss $L_s$, the average gradient with respect to a the k-th entry $b_k$ takes the form of 
\begin{align}
\frac{\partial L_s}{\partial b_k} = \frac{1}{N_s}\sum_{i=1}^{N_s} \mathbbm{1}[y_{i,k}=y_{i,\text{true}}] - \pi_{i,k},
\end{align}
where $\mathbbm{1}$ is the indicator function and $\pi_{i,k}$ is the softmax probability of class $k$ of datapoint $i$. With a randomly initialized model, these softmax probabilities can be assumed to be uniform, leading to an average gradient of
\begin{align}
\frac{\partial L_s}{\partial b_k} = \frac{N_{s,k}}{N_s} - \frac{1}{N_c} = p(y|s)- \frac{1}{N_c},
\end{align}
where $N_c$ is the number of classes.
Upon sending the updated bias vector $b_s = b - \alpha \frac{\partial L_s}{\partial b_k}$ to the server, it can easily reconstruct the marginal label distribution.

\begin{figure}[!htb]
  \centering
  \includegraphics[width=\textwidth]{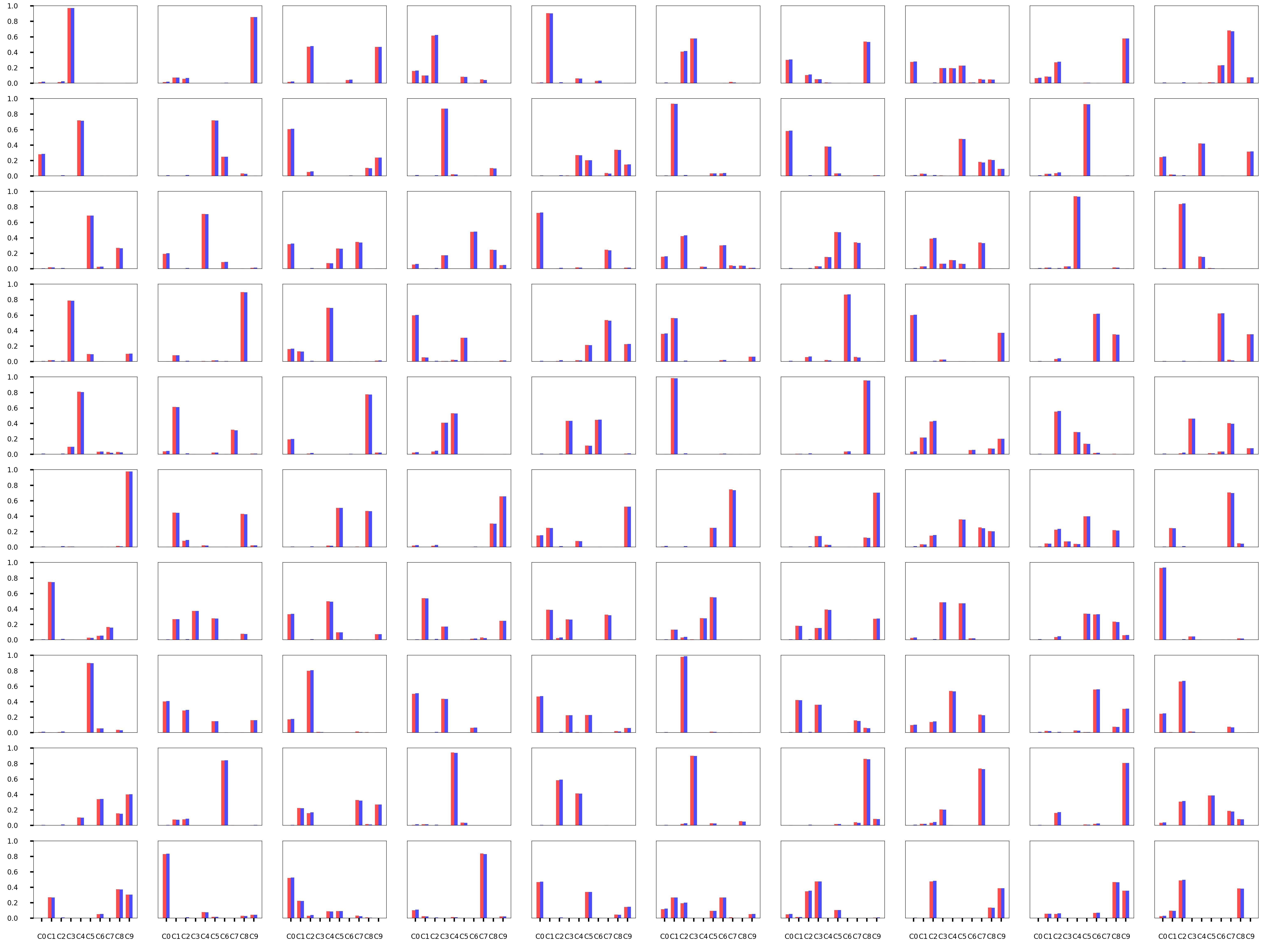}
  \caption{Histogram representation of $p(y|s)$ as well as its server-side reconstruction for every client $s$ for the Cifar10 setup described in Appendix \ref{sec:ExpSetup}. The x-axis per sub-plot enumerates the $10$ classes. For every class $c$, the left bar in red represents $p(y=c|s)$ and the right bar in blue represents its reconstruction. Each client performed a single full data-set update step.}
  \label{fig:priv_single_step}
\end{figure}

\begin{figure}[!htb]
  \centering
  \includegraphics[width=\textwidth]{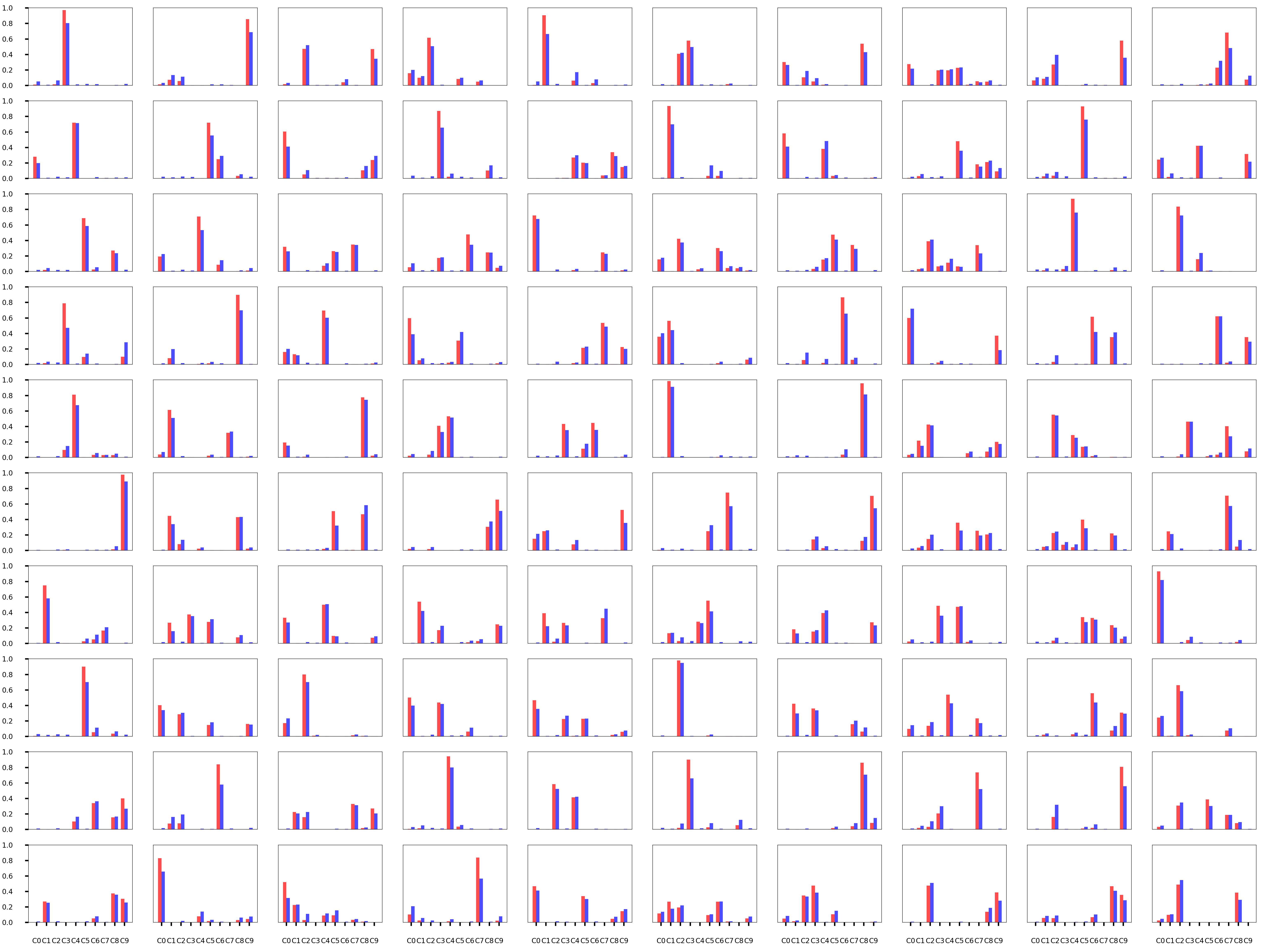}
\caption{Histogram representation of $p(y|s)$ as well as its server-side reconstruction for every client $s$ for the Cifar10 setup described in Appendix \ref{sec:ExpSetup}. The x-axis per sub-plot enumerates the $10$ classes. For every class $c$, the left bar in red represents $p(y=c|s)$ and the right bar in blue represents its reconstruction. Each client performed multiple mini-batch update steps (on average $8$).}
  \label{fig:priv_multi_step}
\end{figure}

Figure \ref{fig:priv_single_step} shows, for every client in our Cifar10 setup (Appendix \ref{sec:ExpSetup}), in red the true marginal $p(y|s)$ and in blue the reconstructed marginal based on (the same) randomly initialized model being sent to each client. Clearly, we have high congruence. In Figure \ref{fig:priv_multi_step}, we investigate the more realistic setup of $E=1$ with mini-batch stochastic gradient descent at the client level. In order to avoid reconstructing the multi-step update, we simply normalize the difference $(b-b_s)$ and interpret it as marginal label distribution. We see that multiple updates (on average: $8$) reduce the congruence between the true and reconstructed marginal, however the information leakage is still remarkable.

\section{Gradient Divergence}
\label{sec:grad_alignment}

We aim to track the divergence of updates for a subset $S'$ of shards at the server at time step $t$ for \fedmix{} and \fedavg{}. Therefore we define a metric inspired by \cite{kim2020on,sattler2019clustered} to define divergence of gradients $\Delta_{k,i}^t = \bs{\omega}_k^t - \bs{\omega}_{k,i}^{t+1}$ for some subset $\bs{\omega}_k$ of the parameters $\bv{w}_k$ of expert $k$ as
\begin{align}
    \text{GD}(\Delta_k^t) = \sum_{i=1}^{S'}\sum_{j=1}^{S'} p(s=i|z=k)p(s=j|z=k) \cdot 0.5 \cdot\left(1-\frac{\Delta_{k,i}^t\cdot \Delta_{k,j}^t}{||\Delta_{k,j}^t||\cdot ||\Delta_{k,i}^t||} \right).
\end{align}

For \fedavg{}, the above metric collapses to
\begin{align}
    \text{GD}(\Delta^t) = \sum_{i=1}^{S'}\sum_{j=1}^{S'} p(s=i)p(s=j) \cdot 0.5\cdot\left(1-\frac{\Delta_{i}^t\cdot \Delta_{j}^t}{||\Delta_{j}^t||\cdot ||\Delta_{i}^t||} \right).
\end{align}

In Figure \ref{fig:GradDivergence} in the main text, we plot the sum of $\text{GD}(\Delta_k^t)$ across all parameters $\bs{\omega}_k$ (\emph{i.e.}, convolutional kernels, weights and biases) of the LeNet-5 experts in comparison to $\bs{\omega}$ for \fedavg{}.

\section{Ablation studies}
\label{sec:ablation}
\begin{figure}[!htb]
\centering
 \begin{subfigure}[b]{0.45\linewidth}
  \centering
  \includegraphics[width=\linewidth]{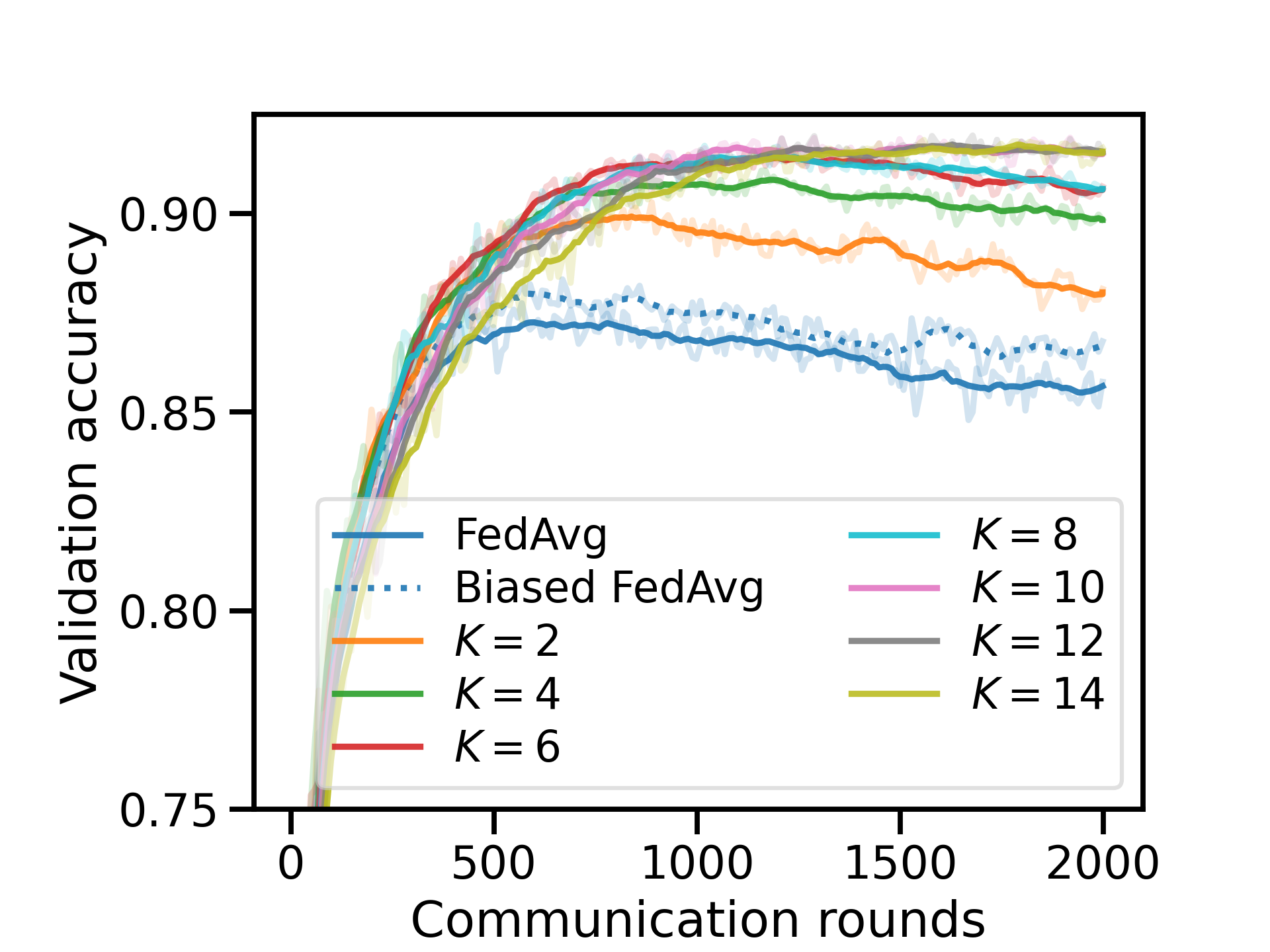}
  \caption{Cifar10: Local validation accuracy}
 \end{subfigure}%
 ~
  \begin{subfigure}[b]{0.45\linewidth}
 \centering
  \includegraphics[width=\linewidth]{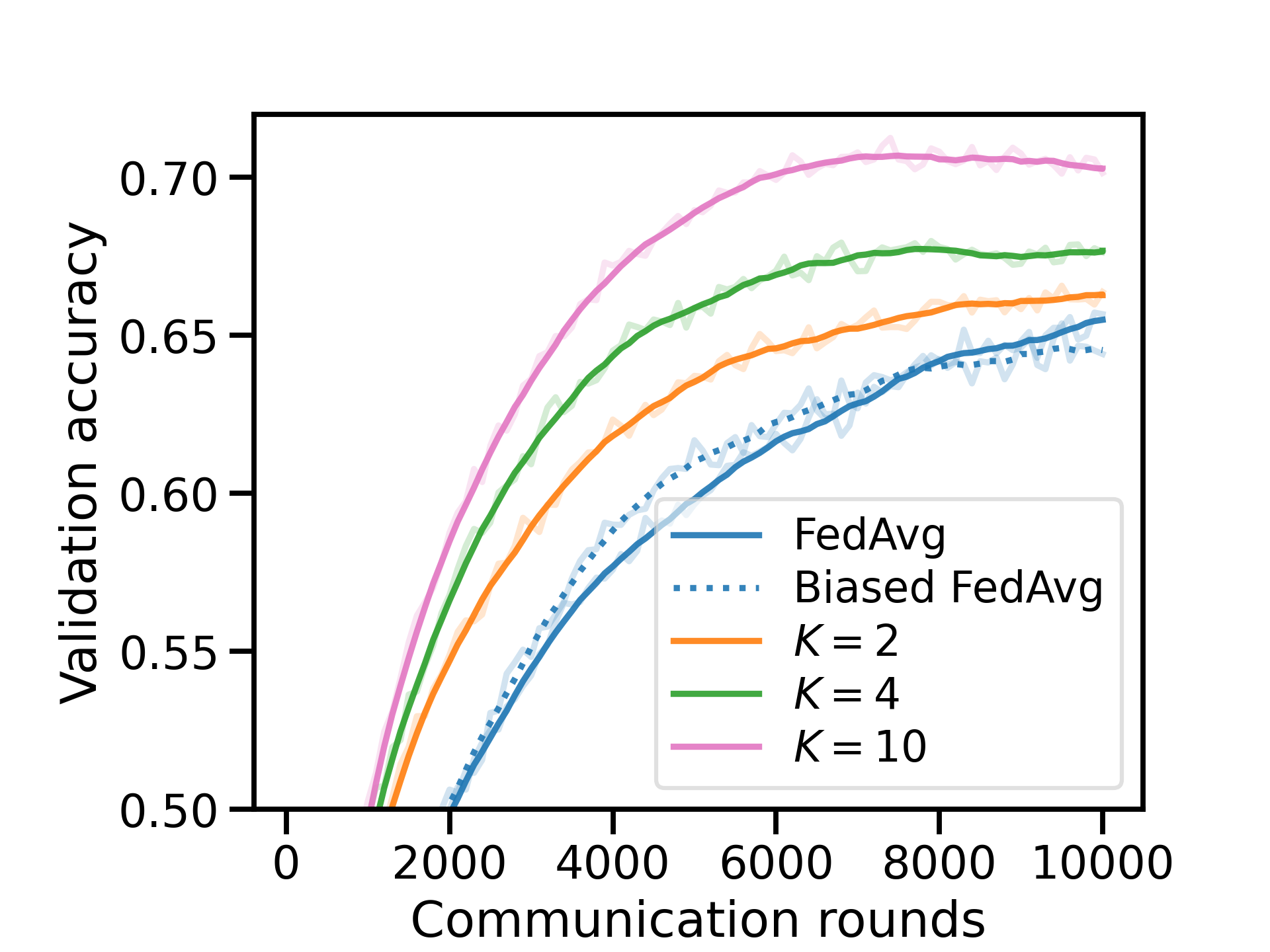}
  \caption{Cifar100: Local validation accuracy}
 \end{subfigure}
 
 \begin{subfigure}[b]{0.45\linewidth}
  \centering
  \includegraphics[width=\linewidth]{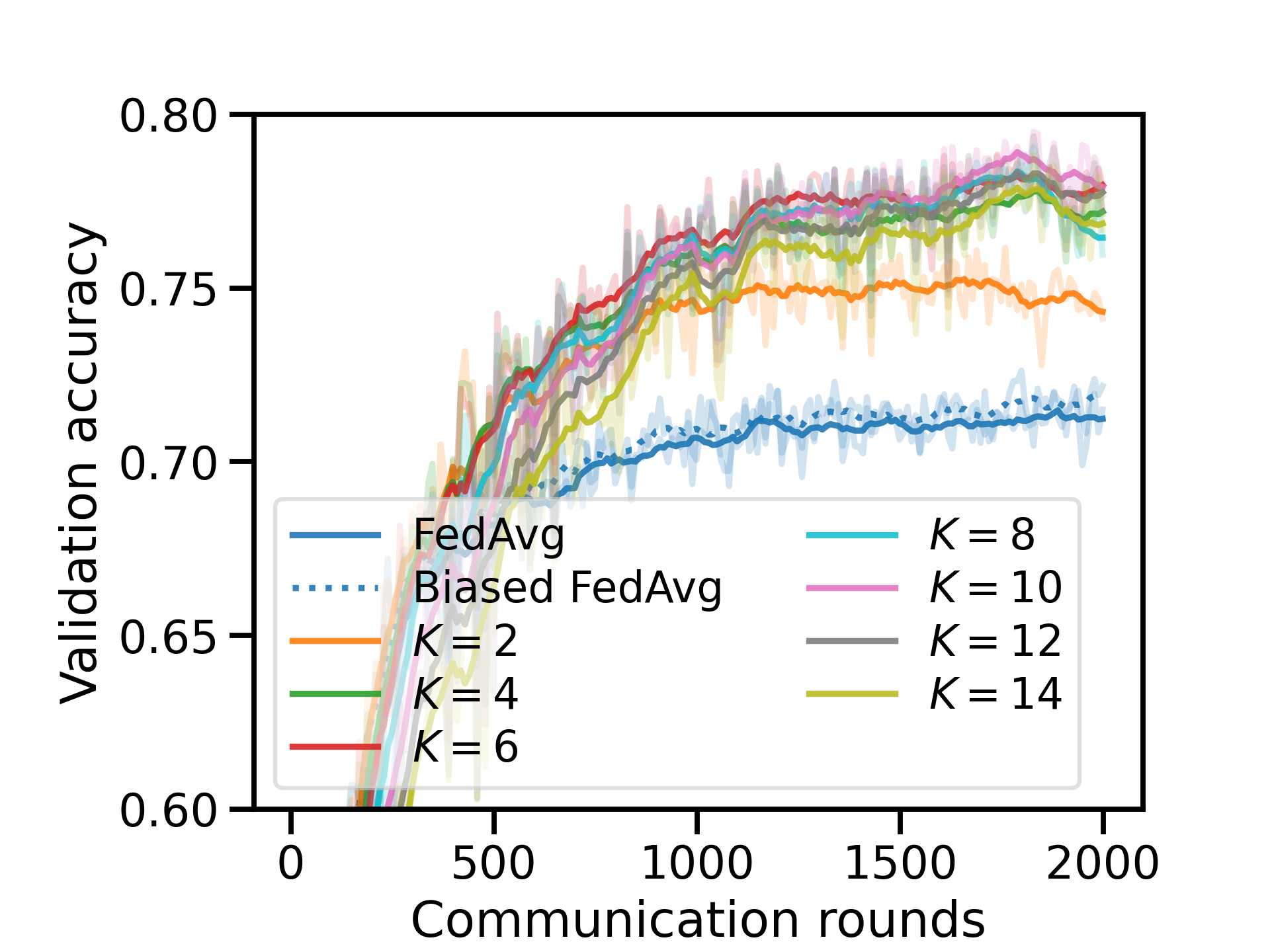}
  \caption{Cifar10: Global validation accuracy}
 \end{subfigure}
 ~
  \begin{subfigure}[b]{0.45\linewidth}
 \centering
  \includegraphics[width=\linewidth]{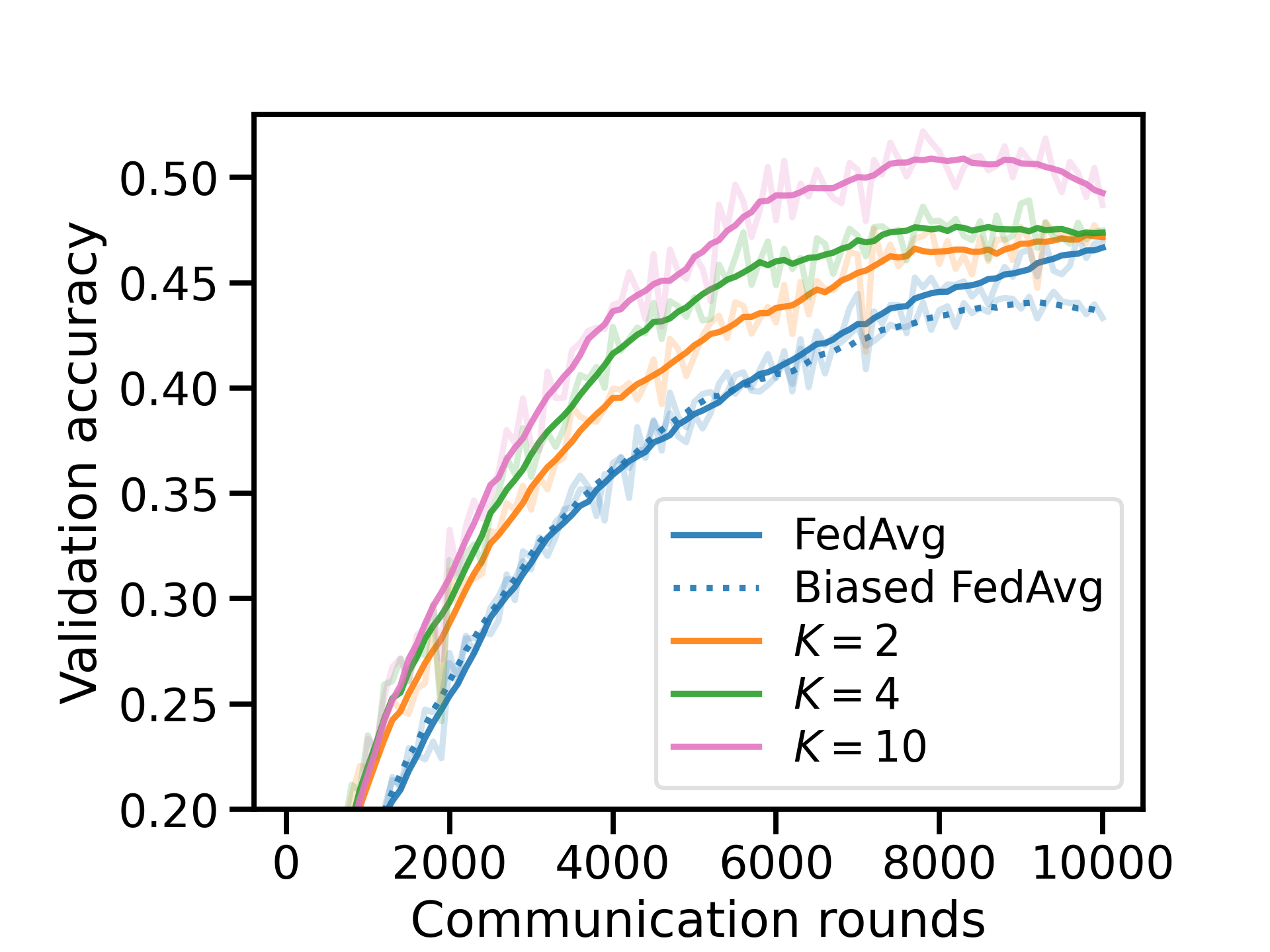}
  \caption{Cifar100: Global validation accuracy}
 \end{subfigure}
 \caption{Ablation studies on the effect of $K$ on the average local accuracy (\textbf{a}, \textbf{b}) and new shard accuracy (\textbf{c},\textbf{d}) on Cifar10 and Cifar100.}
 \label{fig:res_ablation}
\end{figure}

We investigate the characteristics of \fedmix{} by varying the number of experts $K$ for Cifar10 and Cifar100. Figure \ref{fig:res_ablation} shows learning curves of several values of $K$ for these two datasets. We can see that a higher number of experts consistently improves accuracy however with a diminishing rate of return. With increasing $K$, the modeling task for a single expert becomes progressively easier and data across clients becomes more aligned at the cost of higher communication and computation. 

\section{Reducing Communication Costs}
\label{sec:CommCosts}
\begin{figure}[!htb]
\centering
 \begin{subfigure}[b]{0.45\linewidth}
  \centering
  \includegraphics[width=\linewidth]{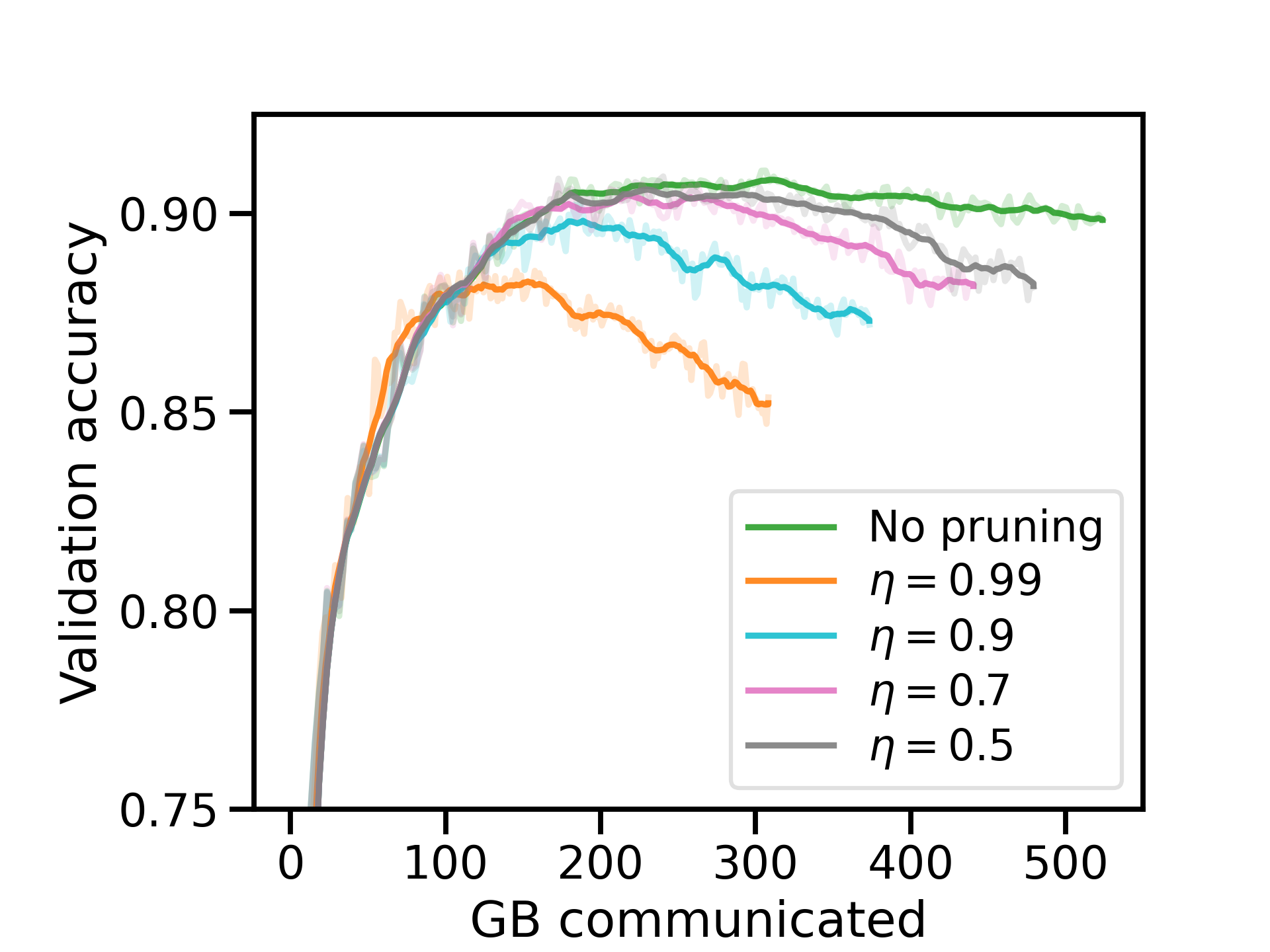}
  \caption{Cifar10: Local validation accuracy}
  \label{fig:res_ablation_KLocal}
 \end{subfigure}%
 ~
  \begin{subfigure}[b]{0.45\linewidth}
 \centering
  \includegraphics[width=\linewidth]{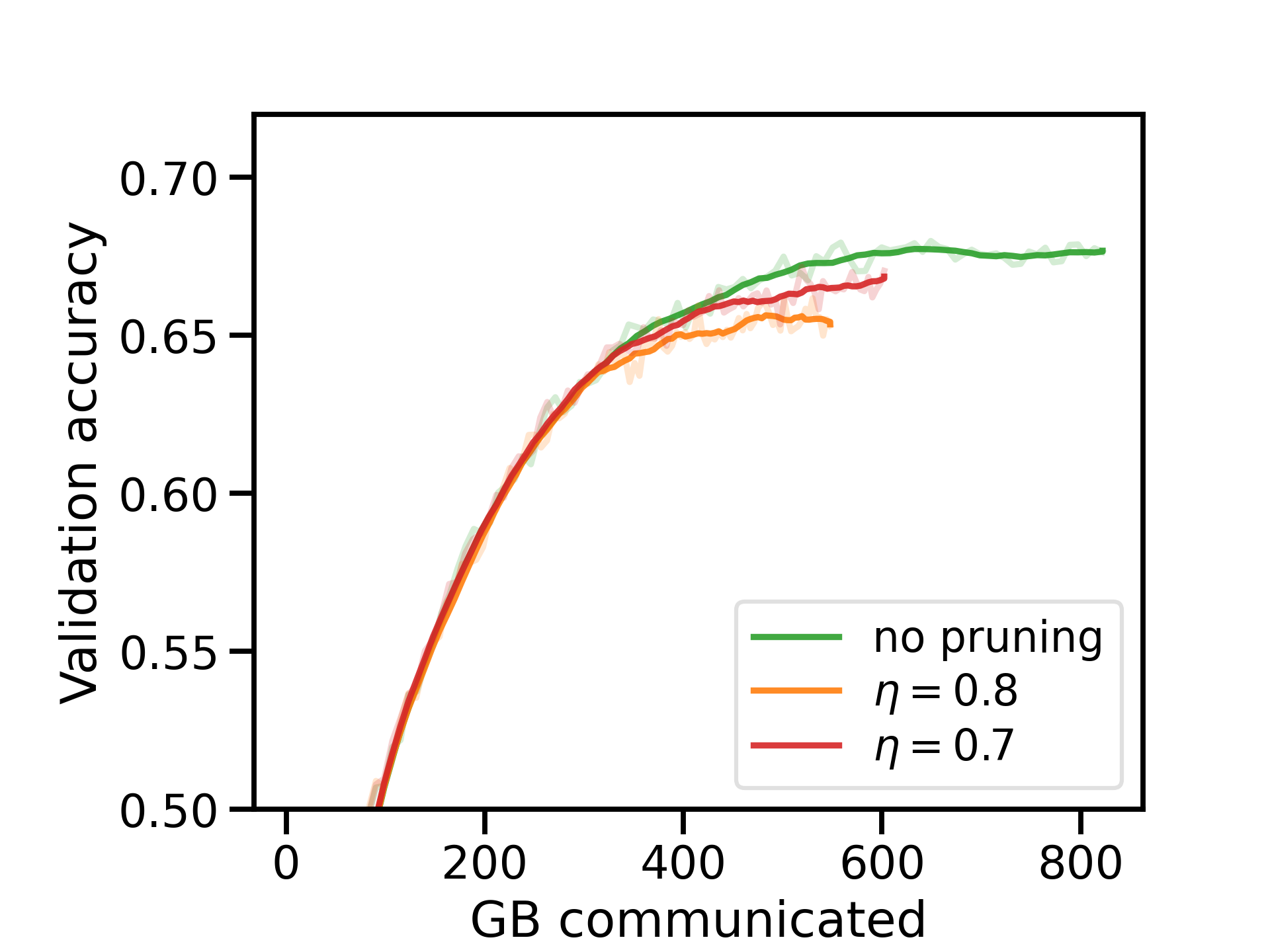}
  \caption{Cifar100: Local validation accuracy}
  \label{fig:res_ablation_pruningLocal}
 \end{subfigure}
 
 \begin{subfigure}[b]{0.45\linewidth}
  \centering
  \includegraphics[width=\linewidth]{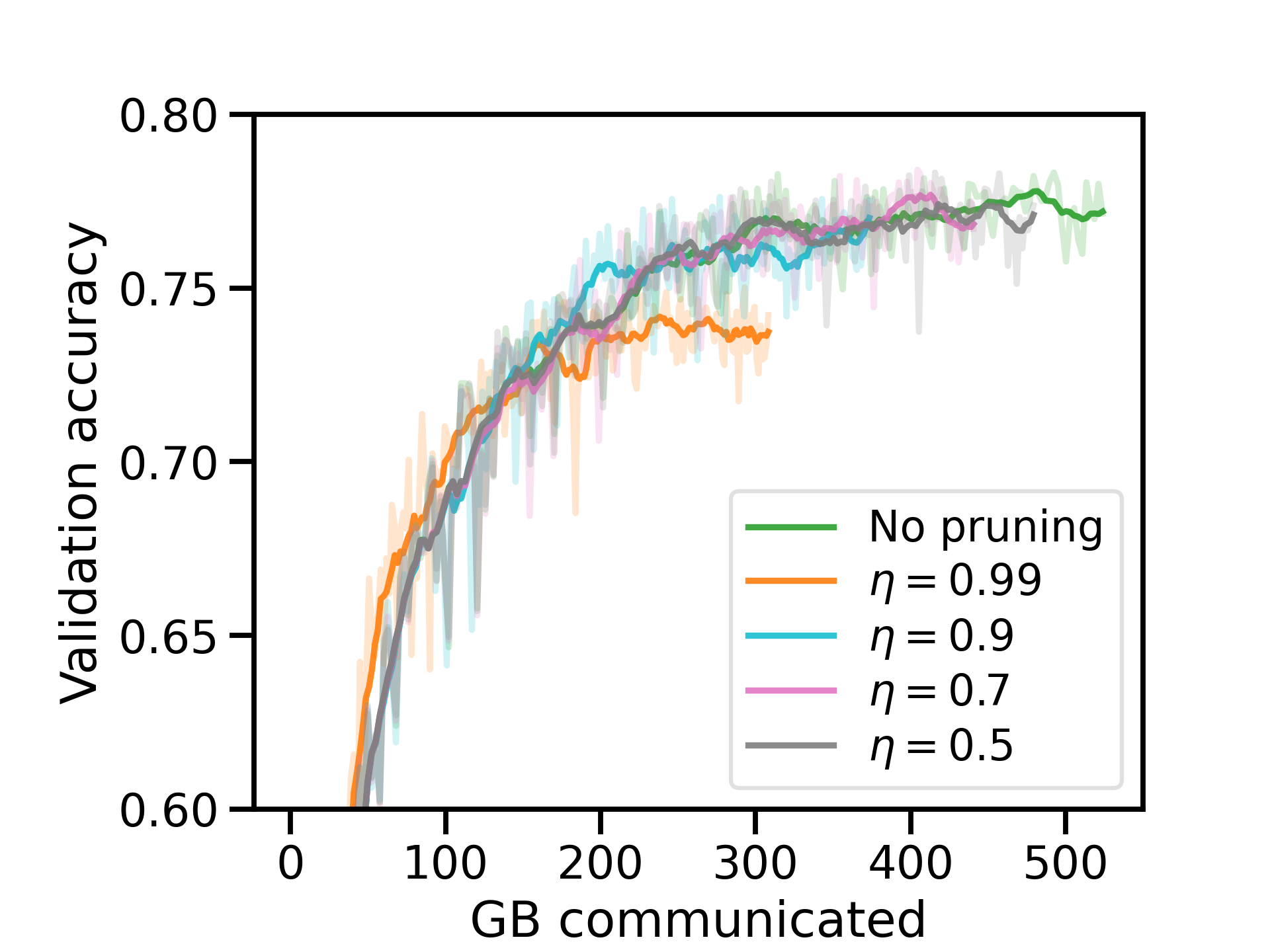}
  \caption{Cifar10: Global validation accuracy}
 \end{subfigure}
 ~
  \begin{subfigure}[b]{0.45\linewidth}
 \centering
  \includegraphics[width=\linewidth]{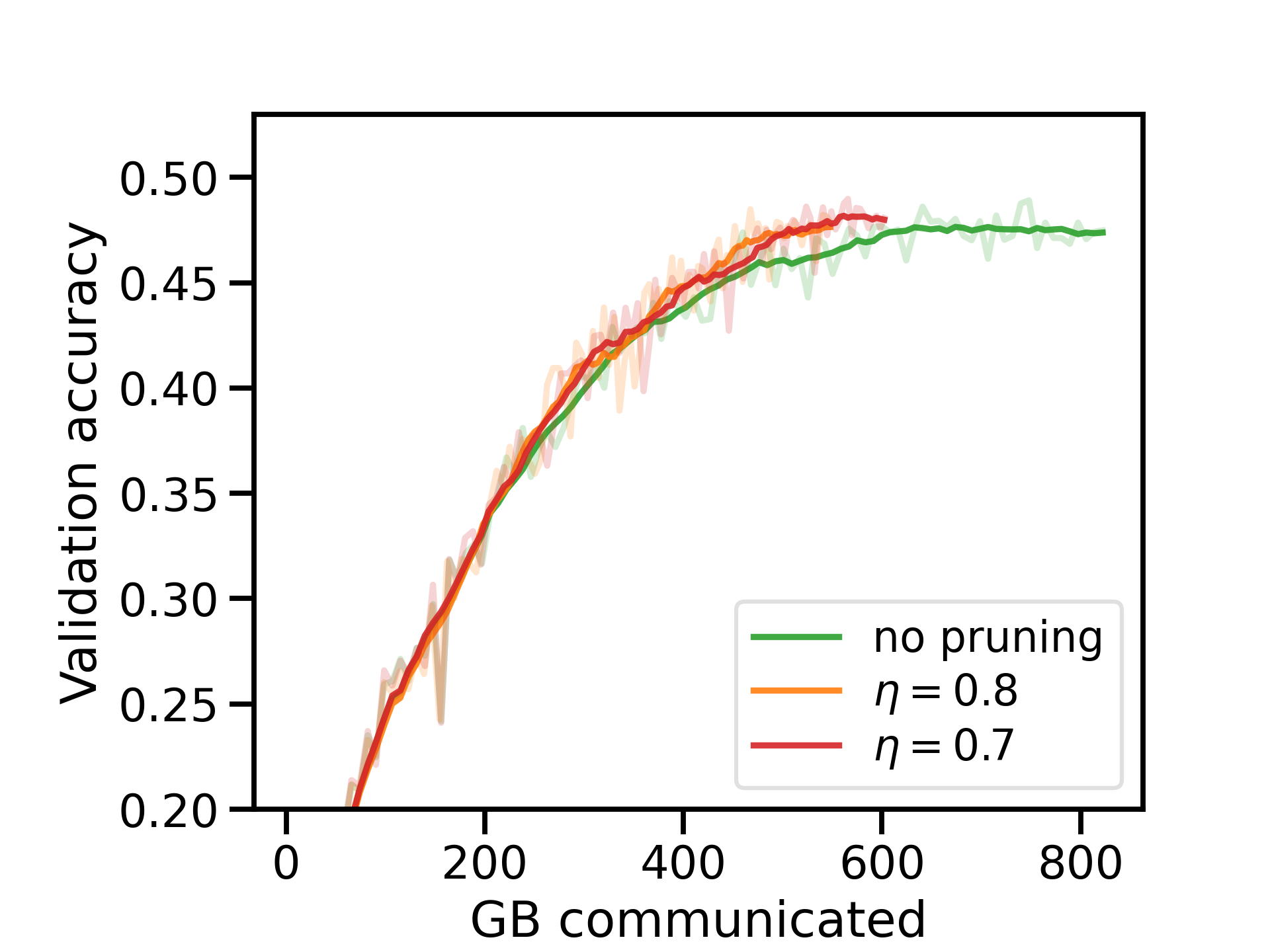}
  \caption{Cifar100: Global validation accuracy}
  \label{fig:res_ablation_pruningGlobal}
 \end{subfigure}
 \caption{The effect of pruning experts on the average local accuracy (\textbf{a}, \textbf{b}) and new shard accuracy (\textbf{c},\textbf{d}) on Cifar10 and Cifar100.}
 \label{fig:res_prune}
\end{figure}

We can ignore the communication of experts between server and client if, for this given client, that particular expert will not be updated during training. If $q_\phi(z=k|s) = E_{p(y|s)}[q_\phi(z=k|y)] = 0$ for an expert $k$ on client $s$, then parameters of that expert observe no gradient and the expert is effectively pruned from the local library of available experts. Since the parameterization of $q_\phi(z|s)$ does not allow for exact values of zero, we introduce $\eta$, such that if $q_\phi(z=k|s)< \eta/K$, we consider expert $k$ to be pruned. Since the value of $q_\phi(z=k|s)$ during local optimization at the client is subject to change, it is possible that $q_\phi(z=k|s)$ briefly lies above the threshold even though it had just been pruned away. Therefore the server considers expert $k$ pruned for a given client $s'$ only if $q_\phi(z=k|s=s')< 0.9 \cdot \eta/K$. Algorithm \ref{alg:fedmix_prune} details this approach. Note that we write $p(z|s)=\mathbb{E}_{\bv{x}\sim D_s}[p_{\theta_s}(z|\bv{x},s)]$, i.e. the true marginal, however in practice we make use of the cheap to compute marginal approximation $q_\phi(z|s) = \mathbb{E}_{y \sim D_s}[q_\phi(z|y)]$. Figure \ref{fig:res_prune} shows experiments with different values of the threshold $\eta$. All models were trained for the same number of communication rounds ($2k$ for Cifar10 and $10k$ for Cifar100). As training progresses and clients begin to drop experts, the amount of GBs communicated is reduced, leading to less GB communicated in total. At the same time, pruning does impact performance significantly, leaving room for improvement in future work. Instead of dropping experts, it might be beneficial to prune weights and features of the individual experts themselves. Inspired by Federated Dropout \citep{caldas2018expanding}, it is interesting to study the effect of dropping experts with a dropout probability related to $q(z|s)$ between communication rounds. 

\begin{algorithm}[!htb]
\caption{The \fedmix{} algorithm. $\alpha, \beta$ are the client and server learning rates; $\gamma$ is a dampening factor and $Z$ a normalization constant. $\eta \in [0,1]$ is the pruning threshold. }\label{alg:fedmix_prune}
\begin{algorithmic}
\Function{Server side}{}
    \State Initialize $\bs{\phi}$ and $K$ vectors $\bv{W} = [\bv{w}_1, \dots, \bv{w}_K]$ 
    \State Initialize $p(z|s)=1/K \forall s$
    \For{round $t$ in $1,\dots T$}
        \State $S' \gets \text{random subset of the clients}$
        \State Initialize $\Delta_{\bv{W}}^t = \mathbf{0}, \Delta_{\bs{\phi}}^t = \mathbf{0}$
        \For{$s$ in $S'$}
            \State $\bv{W}' \gets [\bv{w}_k \,|\, p(z=k|s) \geq 0.9\cdot\eta/K]$ 
            \State $\bv{W}^t_{s}, \bs{\phi}^t_s, p(z|s)\gets$ \Call{Client side}{$s, \bs{\phi}, \bv{W}'$}
            \State Store $p(z|s)$
        \EndFor
        \State $p(s|z) \gets p(z|s) p(s)/\sum_{s\in S'} p(z|s)p(s) $
        \For {$s$ in $S'$}
            \State $\Delta_{\bv{w}_{k}}^t += p(s|z=k) (\bv{w}^{t-1}_{k} - \bv{w}^t_{s, k}) \enskip \forall k$
            \State $\Delta_{\bs{\phi}}^t += \frac{N_s}{N_{S'}}(\bs{\phi}^{t-1} - \bs{\phi}^t_{s})$
        \EndFor
        \State $\Delta_{\bs{\phi}}^t -= \nabla_{\bs{\phi}}H(\sum_c\!q_\phi(z|y\!=\!c)p(y\!=\!c))$
        \State $\bv{w}^{t+1}_{1:K} \gets$ \Call{Adam}{$\Delta_{\bv{w}_{1:K}}^t, \beta$}
        \State $\bs{\phi}^{t+1} \gets$ \Call{Adam}{$\Delta_{\bs{\phi}}^t, \beta$}
    \EndFor
\EndFunction
\\
\Function{Client side}{$s, \bs{\phi}, \bv{W}$}
    \State Get local parameters $\theta_s$
    \For{epoch $e$ in $1, \dots, E$}
        \For {batch $b \in B$}
            \State $q(z|s) \gets \mathbb{E}_{y\sim \mathcal{D}_s}[q_\phi(z|y)]$ 
            \State $K' \gets [k \,|\, q(z=k|s) \geq \eta/K]$ \Comment{Indices of remaining experts}
            \State $\bar{\phi}_c = p_{\bv{w}_z,\theta_s}(y_b=c,z|\bv{x}_b,s)^{1/(\beta N_{s,c})}/Z$ \Comment{Update for remaining experts}
            \State $\phi_c' \gets \bar{\phi}_c \cdot \sum_{k \in K'}\phi_{c,k}$ \Comment{Reweigh to account for pruned experts}
            \State $\phi_c \gets \gamma \phi_c + (1-\gamma)\phi_c' $ \Comment{Dampening and update of $\phi_c$ for the remaining experts}
            \State $\phi_{c,k}'' \gets \frac{\phi_{c,k}}{\sum_{k\in K'}\phi_{c,k}}, k \in K'$ \Comment{Renormalize updated $\phi$ for remaining experts}
            \State $L_s \gets \mathbb{E}_{q_{\phi''}(z|y_b)}[\log  p_{\bv{w}_z}(y_b|\bv{x}_b,z)] + \mathbb{E}_{q_\phi(z|y_b)}[\log p_{\theta_s}(z|\bv{x}_b,s)]$
            \State $\bv{W} += \alpha \nabla_{\bv{W}}L_s$
            \State $\theta_s += \alpha \nabla_{\theta_s} L_s$
        \EndFor
    \EndFor
    \State $q(z|s) \gets \mathbb{E}_{y\sim \mathcal{D}_s}[q_\phi(z|y)]$ 
    \State $\bv{W}' \gets [\bv{w}_k | q(z=k|s) \geq \eta/K]$ \\
    \enskip\enskip\enskip\Return $\bv{W}'; \bs{\phi}; q(z|s)$
\EndFunction
\end{algorithmic}
\end{algorithm}

\begin{figure}[!htb]
\centering
 \begin{subfigure}[b]{0.32\linewidth}
  \centering
  \includegraphics[width=\linewidth]{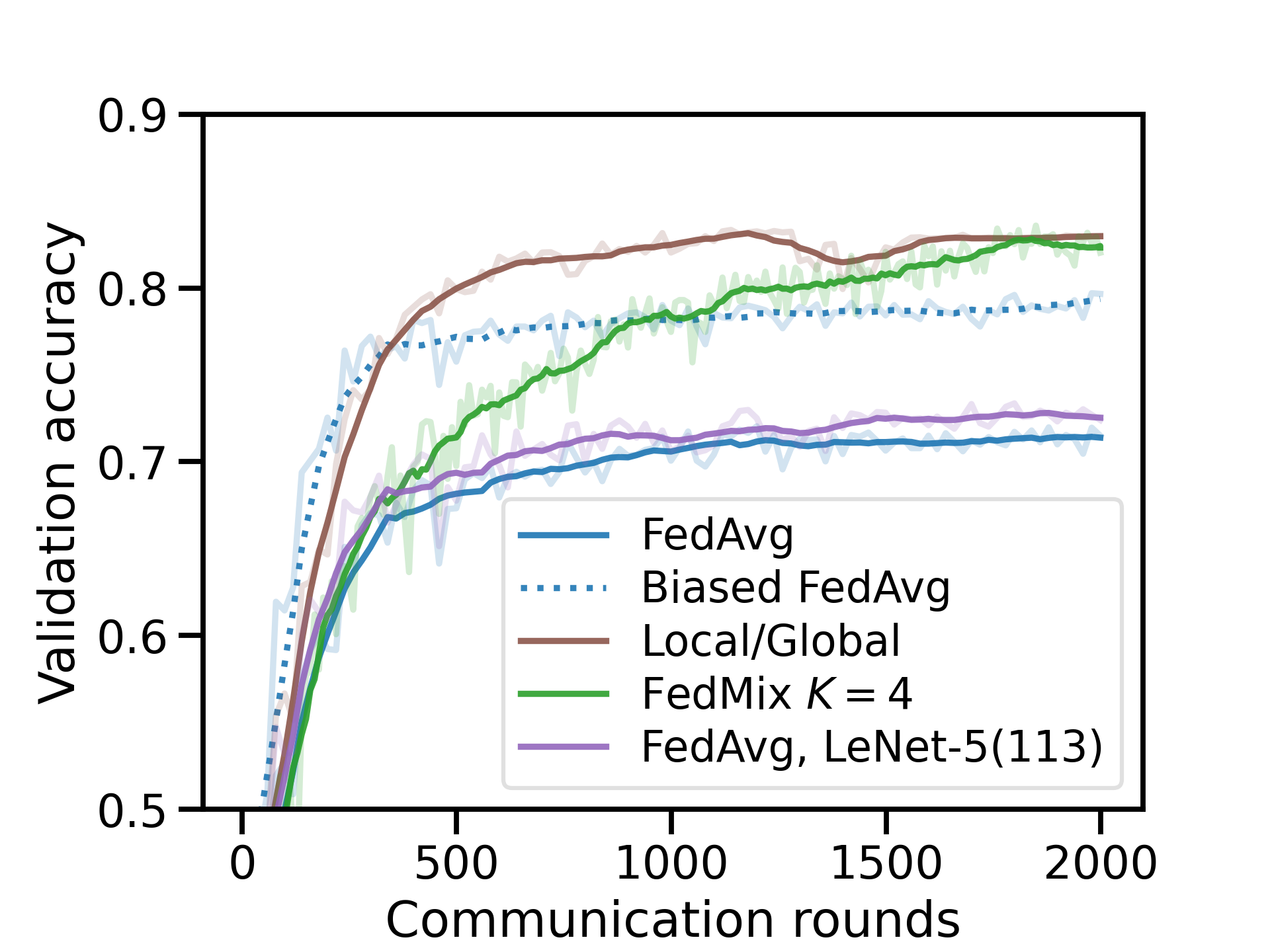}
  \caption{Cifar10}
 \end{subfigure}%
 ~
 \begin{subfigure}[b]{0.32\linewidth}
 \centering
  \includegraphics[width=\linewidth]{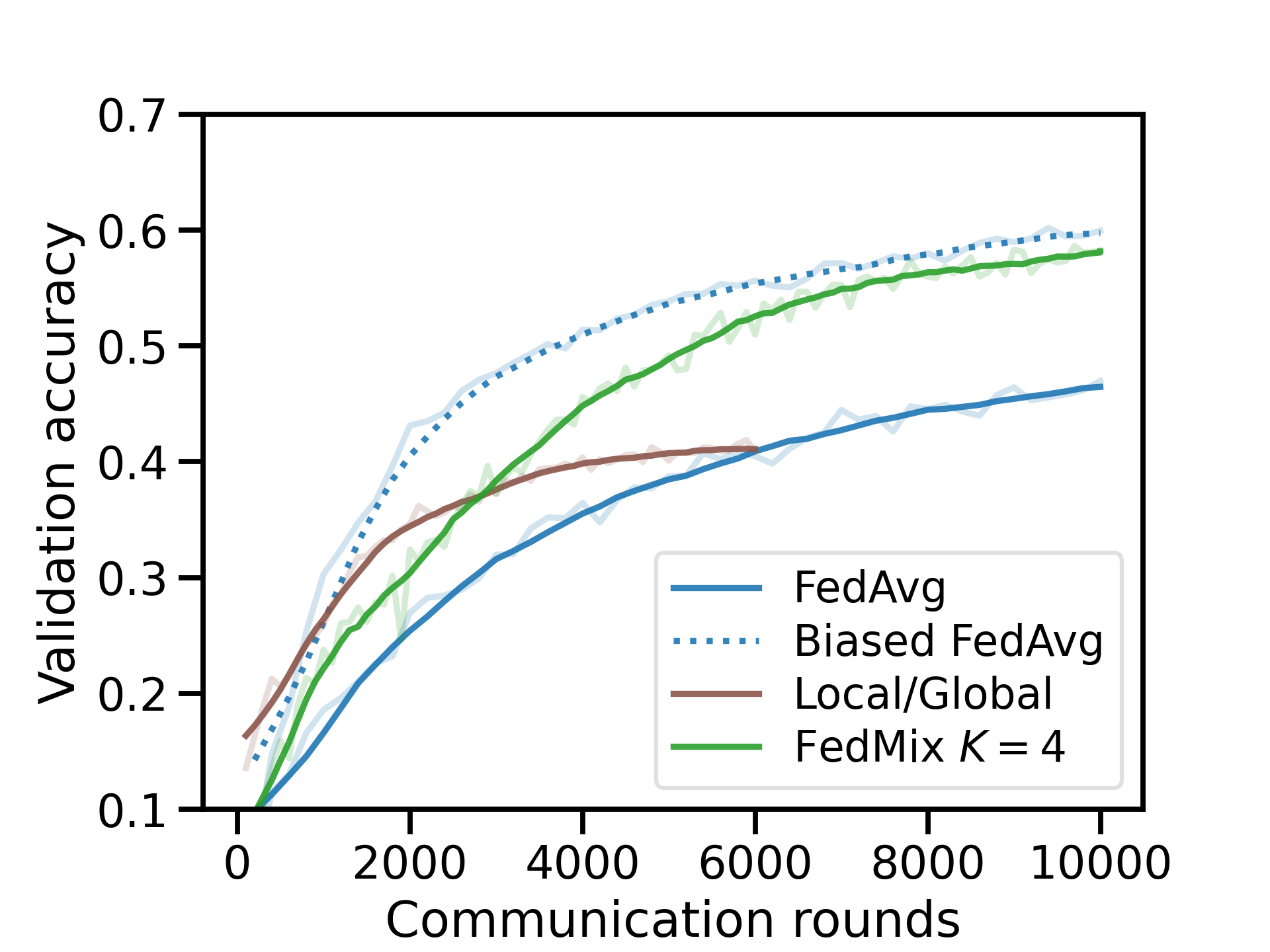}
  \caption{Cifar100}
 \end{subfigure}%
 ~
 \begin{subfigure}[b]{0.32\linewidth}
 \centering
  \includegraphics[width=\linewidth]{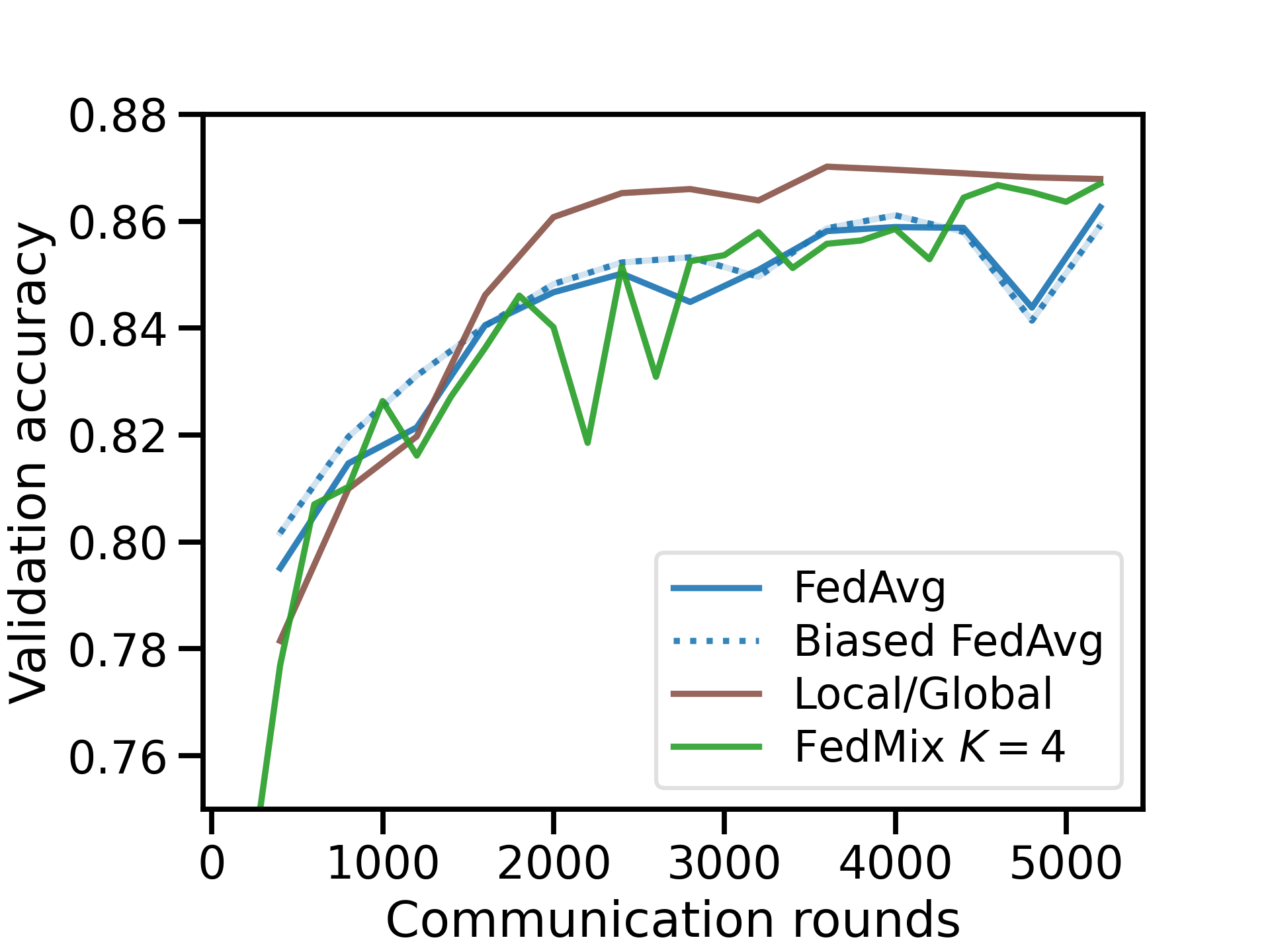}
  \caption{Femnist}
 \end{subfigure}
 
  \caption{Accuracy using the most recent server weights (y-axis) as a function of communication rounds. Cifar10 models are trained on the standard 45k training split. Best viewed in color.}
 \label{fig:server_w_learning_curves}
\end{figure}

\section{Non-Personalized Evaluation}
\label{sec:server_w}

In the main text we discussed how \fedmix{} creates experts that serve as better initialization points for local finetuning. In Figure \ref{fig:server_w_learning_curves} we show learning curves using the server-side model parameters without any local finetuning. We see how \fedmix{} not only serves as a better model for finetuning, but also how the combination of server-side model with the local gating mechanism positions \fedmix{} above \fedavg{}. The server-side model in combination with the locally trained output bias of \textit{biased} \fedavg{} shows the importance of the output bias for combating label skew, however comparing to the results from the main text it seems that standard \fedavg{} can recover the same advantage through finetuning. Local/Global equally shows the strength of combining local feature extractors with the server-side model weights, however cannot recover additional gains through fine-tuning the server-side models.

\section{New Shard Inference}
\label{sec:NewShardInf}
In the main text, we discussed performance of the individual algorithms when evaluating new data on a shard that took part in training and therefore has access to trained local model parameters. This is the case for all considered methods except \fedavg{}, for which there exist no local parameters. Consequently, \fedavg{} can easily be applied to new clients without any local training data. Methods with localized parameters, i.e. \textit{biased} \fedavg{}, Local/Global and \fedmix{} require special consideration.

\cite{liang2020think} propose to ensemble the representations of the local feature extractors across clients before evaluating the global part of the network. We find this approach to work quite poorly in practice. 
In \textit{biased} \fedavg{}, we ensemble the individual local biases across clients to receive a single global bias. For \fedmix{}, we marginalize the local gating predictions to achieve a global gating prediction $p(z|\bv{x}^*) = \sum_{s=1}^S p_{\theta_s}(z|\bv{x}^*,s)p(s)$. Predictions can then be made by marginalizing across experts using this global gating function: $p(y|\bv{x}^*) = \sum_{z=1}^K p_{\bv{w}_k}(y|\bv{x}^*,z)p(z|\bv{x}^*)$. We realize that this is not a scalable solution to the cross-device FL setting, since evaluating $S$ gating functions is costly, however it is feasible in the cross-silo setting. We leave techniques for model distillation for future work.

\begin{figure}[!htb]
\centering
 \begin{subfigure}[b]{0.49\linewidth}
  \centering
  \includegraphics[width=\linewidth]{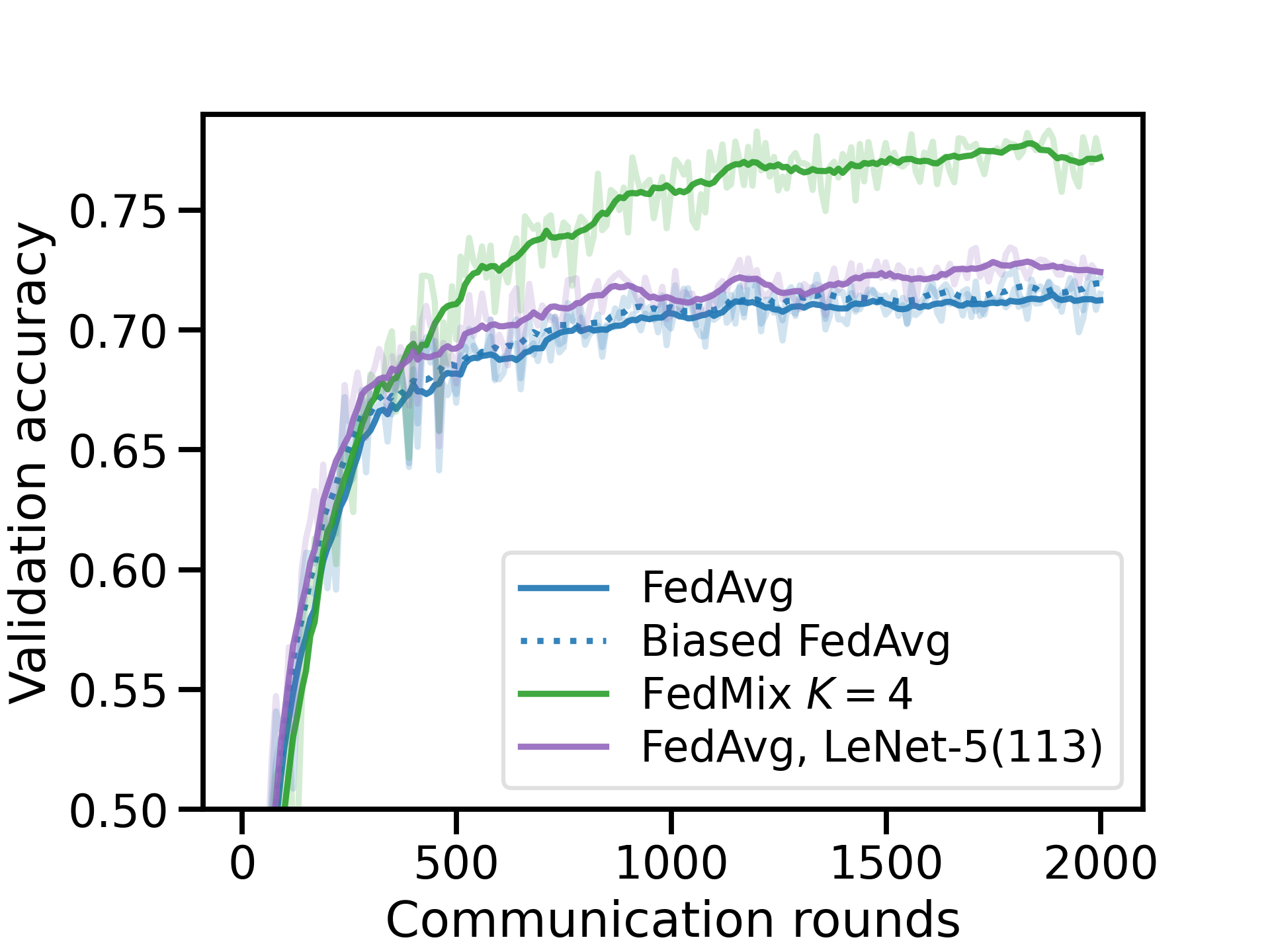}
  \caption{Cifar10}
 \end{subfigure}%
 ~
 \begin{subfigure}[b]{0.49\linewidth}
 \centering
  \includegraphics[width=\linewidth]{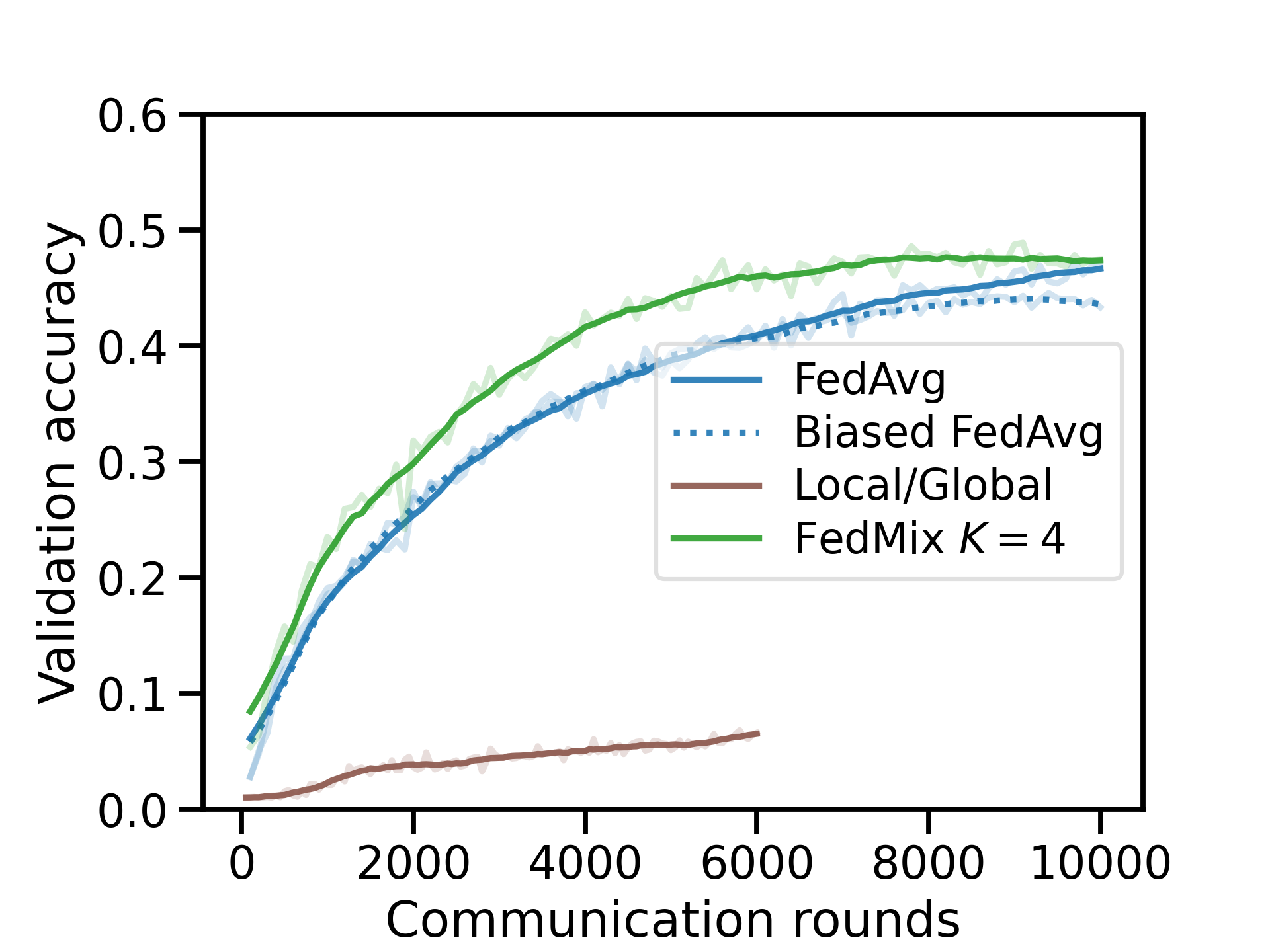}
  \caption{Cifar100}
 \end{subfigure}%
 ~
  \caption{Accuracy on a new client (y-axis) as a function of the amount of communication rounds. Cifar 10 models are trained on the standard 45k training split. Best viewed in color.}
 \label{fig:res_acc_vs_comm_global}
\end{figure}

In Figure \ref{fig:res_acc_vs_comm_global} we use the sum of all local validation sets as a proxy for inference on a new client. Local/Global is left out of the Cifar10 plot since it has random performance throughout. \fedmix{} displays very good performance compared to (\textit{biased}) \fedavg{}, however \fedavg{} eventually catches up with \fedmix{} on Cifar100. In Figure \ref{fig:res_ablation} we see that the global validation accuracy scales with the number of experts $K$.

\subsection{Using a Generative Model}

\begin{wrapfigure}[18]{R}{0.35\textwidth} 
    \centering
    \includegraphics[width=0.24\textwidth]{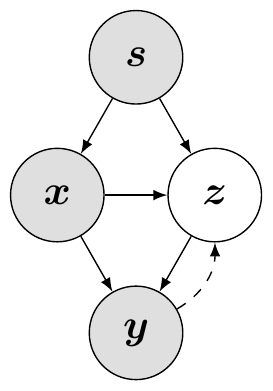}
    \caption{\fedmix{} graphical model. The generative model is depicted with solid lines and the inference model with dashed lines.}
    \label{fig:FedMix_GraphicalMOdel}
\end{wrapfigure}

As an alternative to marginalizing the local gate predictions using $p(s)$, investigating the graphical model in Figure \ref{fig:FedMix_GraphicalMOdel} reveals the possibility for marginalization with $p(s|\bv{x}^*)$:
\begin{align}
    p(z|\bv{x}^*) = \sum_{s=1}^S p_{\theta_s}(z|\bv{x}^*,s)p(s|\bv{x}^*), \quad 
    p(s|\bv{x}^*) \propto p(\bv{x}^*|s)p(s).
\end{align}
It is for this third evaluation case that training local generative models $p(\bv{x}|s)$ for each client becomes interesting, as they allow to compute the responsibilities $p(s|\bv{x}^*)$ at test time. 
In practice, however, the success of this approach depends heavily on the correctness of $p(s|\bv{x}^*)$, which in turn depends on the ability of the local generative models $p(\bv{x}|s)$ to assign high probability to data that resembles $\mathcal{D}_s$ and low probability to out-of-distribution data. Training and calibrating generative models for this task is in itself an active area of research \citep{nalisnick2018deep} and investigating how clients in a federated setting might exchange information to facilitate this process is not yet explored. We therefore leave a thorough evaluation of this case to future work and only present here a MNIST \citep{lecun2010mnist} experiment.

We train \fedmix{} with $K=4$ and experts of two hidden layer ReLU MLP with 200 hidden units on MNIST. We split the dataset into $S=100$ clients according to the procedure described in \cite{liang2020think}. \fedmix{} achieves $97.7\%$ average validation accuracy compared to $97.0\%$ with \fedavg{} after $600$ communication steps. Independently for each client, we train a small variational autoencoder with a $32$-dimensional latent space using a two-layer MLP with $512$ and $256$ hidden units respectively as encoder and mirrored decoder structure. We optimize the VAEs using Adam with standard hyper parameters, $B=10$ and perform early stopping on the local validation sets after no improvement for three epochs. After training, each client communicates their VAE to the server, where we evaluate $p(s|\bv{x}^*)$ to marginalize over the local gating functions according to $p(z|\bv{x}^*)=\sum_{s=1}^S p(z|\bv{x}^*,s)p(s|\bv{x}^*)$. With this procedure, \fedmix{} achieves $95.9\%$ test set accuracy, compared to \fedavg{} with $96.9\%$. Marginalizing with $p(s)$ instead of $p(s|\bv{x}^*)$ achieves $96.63\%$, showing the limitation of the approach in that any error in $p(s|\bv{x}^*)$ propagates into the expert assignment. Reliable out-of-distribution detection capabilities in the individual estimators for $p(\bv{x}|s)$ are therefore necessary.


\end{document}